\documentclass[10pt,journal,compsoc]{IEEEtran}
\ifCLASSOPTIONcompsoc
  \usepackage[nocompress]{cite}
\else
  \usepackage{cite}
\fi
\ifCLASSINFOpdf
\else
\fi

\usepackage{graphicx}
\usepackage{amsmath}
\usepackage{amssymb}
\usepackage{booktabs}
\usepackage{xcolor}
\newcommand{\bp}[1]{\textcolor{black}{#1}}

\makeatletter
\renewcommand{\maketag@@@}[1]{\hbox{\m@th\normalsize\normalfont#1}}%
\makeatother
\usepackage{algorithm}  
\usepackage{algorithmic} 
\usepackage[colorlinks,linkcolor=red]{hyperref}
\usepackage[accsupp]{axessibility}
\usepackage{ragged2e}
\usepackage{bbding}
\usepackage{multirow}
\usepackage{subcaption}
\begin{document}
%
\title{$\beta$-DARTS++: Bi-level Regularization for Proxy-robust Differentiable Architecture Search}
\author{
Peng Ye, Tong He, Baopu Li, Tao Chen, Lei Bai, Wanli Ouyang
\IEEEcompsocitemizethanks{\IEEEcompsocthanksitem Peng Ye and Tao Chen are with the School of Information Science and Technology, Fudan University, Shanghai, China \\
E-mail: \{20110720039, eetchen\}@fudan.edu.cn
\IEEEcompsocthanksitem Tong He, Lei Bai and Wanli Ouyang are with the Shanghai AI Laboratory, Shanghai, China.
\IEEEcompsocthanksitem Baopu Li is with the Oracle Health and AI, Oracle, USA.
\IEEEcompsocthanksitem Work is partially performed when Peng Ye is an intern at Shanghai AI Laboratory, China. Tao Chen is the corresponding author.}
}

\IEEEtitleabstractindextext{%
\begin{abstract}\justifying
Neural Architecture Search~(NAS) has attracted increasing attention in recent years because of its capability to design neural networks automatically. Among them, differential NAS approaches such as DARTS, have gained popularity for the search efficiency. However, they still suffer from three main issues, that are, the weak stability due to the performance collapse, the poor generalization ability of the searched architectures, and the inferior robustness to different kinds of proxies (i.e., computationally reduced settings). To solve the stability and generalization problems, a simple-but-effective regularization method, termed as Beta-Decay, is proposed to regularize the DARTS-based NAS searching process (referred as $\beta$-DARTS). Specifically, Beta-Decay regularization can impose constraints to keep the value and variance of activated architecture parameters from being too large, thereby ensuring the fair competition among architecture parameters and making the supernet less sensitive to the impact of input on the operation set. In-depth theoretical analyses on how it works and why it works are provided. Comprehensive experiments on a variety of search spaces and datasets validate that Beta-Decay regularization can help to stabilize the searching process and makes the searched network more transferable across different datasets. 
To address the robustness problem,
we first benchmark different NAS methods 
under a wide range of proxy data, proxy channels, proxy layers and proxy epochs, since the robustness of NAS under different kinds of proxies has not been explored before. 
We then conclude some interesting findings and find that $\beta$-DARTS always achieves the best result among all compared NAS methods under almost all proxies.
We further introduce the novel flooding regularization to the weight optimization of $\beta$-DARTS (termed as Bi-level regularization), and experimentally and theoretically verify its effectiveness for improving the proxy robustness of differentiable NAS. In summary, our search scheme shows lots of outstanding properties for practical applications,
and the code is available at \href{https://github.com/Sunshine-Ye/Beta-DARTS}{https://github.com/Sunshine-Ye/Beta-DARTS}.
\end{abstract}

}

\maketitle

\IEEEdisplaynontitleabstractindextext

%
\IEEEpeerreviewmaketitle

\IEEEraisesectionheading{\section{Introduction}\label{sec:introduction}}

\begin{figure}[t] 
	\centering
	\includegraphics[width=3.2in]{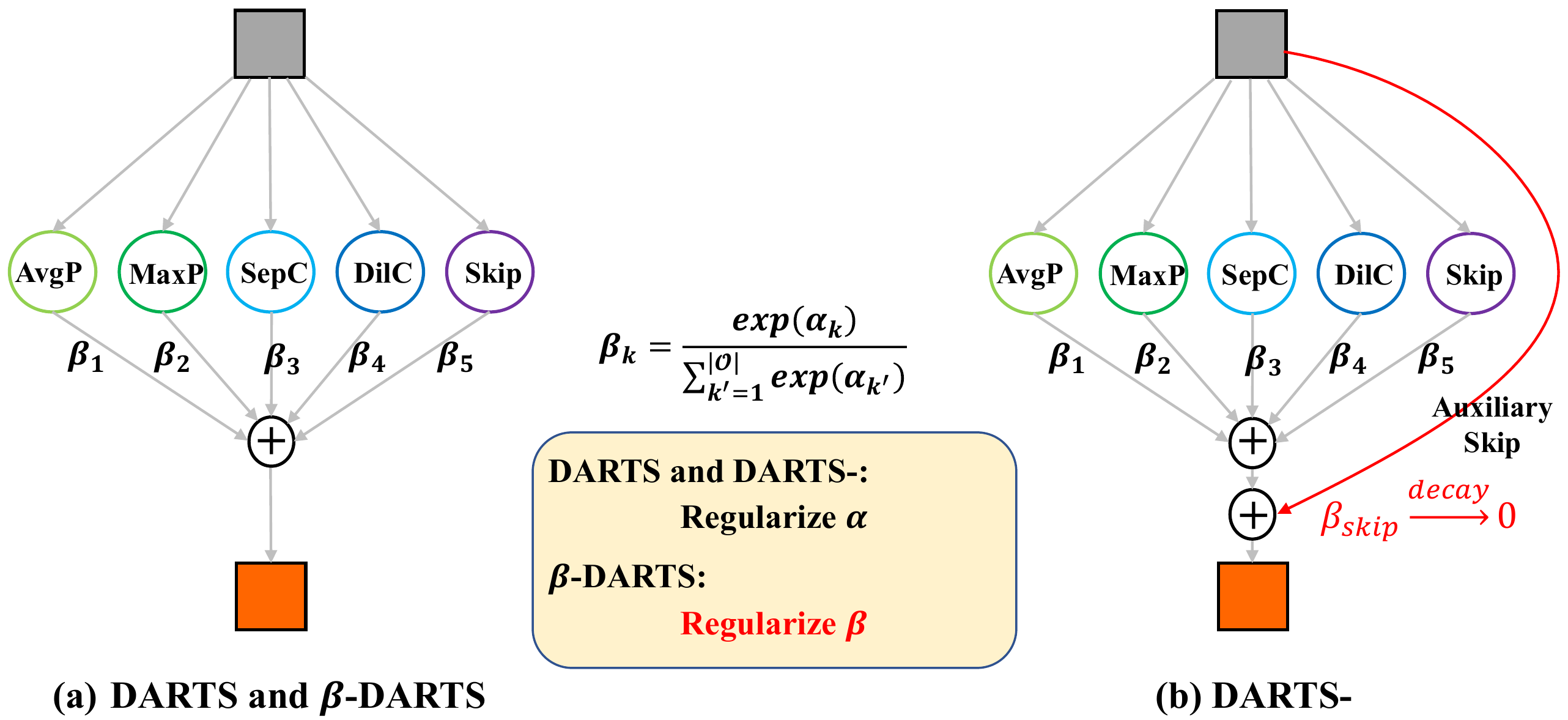}
	\vspace{-6pt}
	\caption{Schematic illustration about (a) DARTS~\cite{darts} and our proposed $\beta$-DARTS, (b) DARTS-~\cite{darts-}.
	DARTS- adds an auxiliary skip connection with a decay rate $\beta_{skip}$
	to alleviate the performance collapse problem. $\beta$-DARTS introduces the Beta-Decay regularization to improve both the stability of the searching process and the generalization ability of the searched architecture. Best viewed in color.}
	\label{fig:1}
	\vspace{-10pt}	
\end{figure}

%
%
%
%
\IEEEPARstart{N}{eural} architecture search (NAS) has attracted lots of interest for its potential to automatize the process of architecture design. Previous reinforcement learning~\cite{nasnet, mnasnet} and evolutionary algorithm~\cite{amoebanet} based methods usually incur massive computation overheads, which hinder their practical applications. To reduce the search cost, a variety of approaches are proposed, including performance estimation~\cite{klein2016learning}, network morphisms~\cite{cai2018path} and one-shot architecture search~\cite{spos,ye2022efficient,ye2022stimulative,darts}. In particular, one-shot methods resort to weight sharing technique, which only needs to train a supernet covering all candidate sub-networks once. Based on this weight sharing strategy, differentiable architecture search~\cite{darts} (namely DARTS, as shown in Fig.~\ref{fig:1}) relaxes the discrete operation selection problem to learn differentiable architecture parameters, which further improves the search efficiency by alternately optimizing supernet weights and architecture parameters.

\begin{table*}[t]
\caption{Comprehensive comparisons of various NAS methods on NAS-Bench-201. $\beta$-DARTS++ equipping with Beta-Decay regularization and Weight Flooding regularization can guarantee both search stability and generalization, and perform more robustly under various proxies. {\scriptsize\CheckmarkBold}/{\scriptsize\XSolidBrush} stands for yes/no. Note that org denotes the original baseline setting.
}
\label{tab:total_comp}
\centering
\setlength{\tabcolsep}{1.3mm}{
\begin{tabular}{lcccccccc}
\hline
Method       & \begin{tabular}[c]{@{}c@{}}Network\\ Weights $w$\end{tabular} & \begin{tabular}[c]{@{}c@{}}Arch \\ Params $\alpha$\end{tabular} & \begin{tabular}[c]{@{}c@{}}Arch \\ Params $\beta$\end{tabular} & \begin{tabular}[c]{@{}c@{}}Stability and \\ Generalization\end{tabular} & \begin{tabular}[c]{@{}c@{}}Proxy\\ Epochs\end{tabular} & \begin{tabular}[c]{@{}c@{}}Proxy\\ Data\end{tabular} & \begin{tabular}[c]{@{}c@{}}Proxy\\ Layers\end{tabular} & \begin{tabular}[c]{@{}c@{}}Proxy\\ Channels\end{tabular} \\ \hline
DARTS~\cite{darts}/SNAS~\cite{snas}/… & L2                                                          & L2                                                                   & \XSolidBrush                                                                    & \XSolidBrush                                                                       & 100                                                    & 100\%                                                & 5                                                      & 16                                                       \\
RDARTS-L2~\cite{rdarts}   & Larger L2                                                   & L2                                                                   & \XSolidBrush                                                                    & \XSolidBrush                                                                       & 100                                                    & 100\%                                                & 5                                                      & 16                                                       \\
SDARTS-RS~\cite{sdarts}    & Random Smoothing                                            & L2                                                                   & \XSolidBrush                                                                    & \XSolidBrush                                                                       & 100                                                    & 100\%                                                & 5                                                      & 16                                                       \\
SDARTS-ADA~\cite{sdarts}   & Adaversarial Training                                       & L2                                                                   & \XSolidBrush                                                                    & \XSolidBrush                                                                       & 100                                                    & 100\%                                                & 5                                                      & 16                                                       \\
EcoNAS~\cite{zhou2020econas}       & L2                                                          & L2                                                                   & \XSolidBrush                                                                    & \XSolidBrush                                                                       & 1/10*org                                               & 100\%                                                & 5                                                      & 1/4*org                                                  \\
Proxy Data~\cite{na2021accelerating}   & L2                                                          & L2                                                                   & \XSolidBrush                                                                    & \XSolidBrush                                                                       & 100                                                    & 10\%/20\%                                            & 5                                                      & 16                                                       \\
PC-DARTS~\cite{pc-darts}     & L2                                                          & L2                                                                   & \XSolidBrush                                                                    & \XSolidBrush                                                                      & 100                                                    & 100\%                                                & 5                                                      & 1/4*org                                          \\
EPC-DARTS~\cite{xu2021partially}     & L2                                                          & L2                                                                   & \XSolidBrush                                                                    & \XSolidBrush                                                                      & 100                                                    & 100\%                                                & 5                                                      & 1/8*org                                          \\
$\beta$-DARTS~\cite{ye2022beta}      & L2                                                          & L2                                                                   & Beta-Decay                                                           & \CheckmarkBold                                                                     & 100                                                    & 100\%                                                & 5                                                      & 16                                                       \\
$\beta$-DARTS++    & Weight Flooding                                       & L2                                                                   & Beta-Decay                                                           & \CheckmarkBold                                                                       & Adaptive                                               & 1\%-100\%                                            & 1-5                                                    & 1-16                                                     \\ \hline
\end{tabular}}
\vspace{-10pt}
\end{table*}

Although the differentiable method has the advantages of simplicity and computational efficiency, \bp{the challenges of} its stability, architecture generalization, and proxy robustness  still need to be fully resolved. 
Firstly, lots of studies have shown that DARTS frequently suffers from performance collapse, that is, the searched architecture tends to \bp{prefer} parameter-free operations especially for skip connection, leading to the performance degradation~\cite{darts-,rdarts}. To handle this stability challenge, lots of \bp{interesting works have been} proposed: directly restricting the number of skip connections~\cite{pdarts,darts+}; exploiting or regularizing relevant indicators such as the norm of Hessian regarding the architecture parameters~\cite{rdarts,sdarts}; changing the searching and/or discretization process~\cite{fairdarts,dots,darts-}; implicitly regularizing the learned architecture parameters~\cite{rdarts}. However, the explicit regularization of architecture parameters optimization receives little attention, as previous works (including the above methods) adopt L2 or weight decay regularization by default on learnable architecture parameters (i.e., $\alpha$), without exploring solutions along this direction. 
Secondly, several works have pointed out that the optimal architecture obtained on the specific dataset cannot guarantee its good performance on another dataset~\cite{adaptNAS,mixsearch}, namely the architecture generalization challenge. To improve the generalization of the searched model, AdaptNAS~\cite{adaptNAS} explicitly minimizes the generalization gap of architectures between domains via the idea of cross-domain. 
MixSearch~\cite{mixsearch} searches a generalizable architecture by mixing multiple datasets of different domains and tasks. 
However, both methods solve this issue by leveraging larger datasets, while how to use a single dataset to learn a generalized architecture remains challenging. 
Thirdly, in order to relieve the unacceptable search cost, proxies (i.e., computationally reduced settings) are widely used in the NAS field. A few recent works further propose some better proxies to realize more efficient or effective NAS. EcoNAS~\cite{zhou2020econas} systematically analyzes the effect of different proxies on evolutionary-based NAS and finds some fast and reliable proxies. PC-DARTS~\cite{pc-darts} and EPC-DARTS~\cite{xu2021partially} explore how to randomly sample a subset of input channels to reduce the memory cost of differentiable NAS. Proxy Data~\cite{na2021accelerating} designs a new proxy data selection method for efficient NAS. However, all these works focus on finding better proxies for NAS, while the robustness of different NAS methods under various proxies has not been explored. This paper is dedicated to simultaneously solving the above-mentioned three challenges in an efficient way. 

Inspired by the widely-used L2~\cite{cortes2012l2} or weight decay regularization~\cite{wd} approaches, we first intend to design a customized regularization for DARTS-based methods, which can explicitly regularize the optimizing process of architecture parameters. However, different from the regularization on the learnable architecture parameter set, $\alpha$ (before the nonlinear activation of softmax), commonly used in standard DARTS and its subsequent variants, we propose a novel and generic Beta-Decay regularization, imposing regularization on the activated architecture parameters $\beta$ (after softmax), where $\beta_k=\frac{\exp \left(\alpha_{k}\right)}{\sum_{k^{\prime}=1}^{\left | \mathcal{O}\right |} \exp \left(\alpha_{k^{\prime}}\right)}$.
On one hand, the proposed Beta-Decay regularization is very simple to implement, achieved with only  \textbf{one additional line of PyTorch code} in DARTS~(Alg~\ref{pytorch_implementation}). On the other hand, this simple implementation is grounded by in-depth theoretical support. We provide theoretical analysis to show that, Beta-decay regularization not only mitigates unfair competition advantage among operations and solves the domination problem of parameter-free operations, but also minimizes the Lipschitz constraint defined by architecture parameters and makes sure the generalization ability of searched architecture. In addition, we mathematically and experimentally demonstrate that, commonly-used L2 or weight decay regularization on $\alpha$ may not be effective or even counterproductive for improving the stability and generalization of DARTS. 
DARTS with Beta-Decay regularization ($\beta$-DARTS) is illustrated in Fig.~\ref{fig:1}. Extensive experiments on various search spaces (i.e. NAS-Bench-201, DARTS, MobileNet and NAS-Bench-1Shot1) and datasets (i.e. CIFAR-10, CIFAR-100, ImageNet16-120 and ImageNet-1K) verify the stability and generalization of the proposed method. 
\vspace{-3pt}
\begin{algorithm}[htb]
\label{alg:beta-decay code}   
\begin{algorithmic}[1] 
\STATE $\mathcal{L}_{Beta}=$  torch.mean(torch.logsumexp( \\
\qquad \quad self.model.\_arch\_parameters, dim=-1))
\STATE loss $=$ self.\_val\_loss(self.model, input\_valid, \\ 
\qquad \quad target\_valid)+$\lambda\mathcal{L}_{Beta} $
\end{algorithmic}  
\caption{PyTorch Implementation in DARTS}
\label{pytorch_implementation}
\end{algorithm} 
\vspace{-3pt}

Since the robustness of different NAS under \bp{various} kinds of proxies has not been investigated before, we first benchmark different NAS methods including our $\beta$-DARTS under a wide range of proxy data, proxy channels, proxy layers and proxy epochs, and conclude some interesting findings, as shown in Sec~\ref{sec: proxy robustness}. For example: 1) reducing the proxy data degrades the performance of different NAS consistently; 2) reducing the proxy channels improves the performance of DARTS/RDARTS-L2/SDARTS-RS to some extent; 3) reducing the proxy layers improves the performance of RDARTS-L2 while degrades the performance of SDARTS-RS; 4) the last epoch is not necessarily the optimal epoch for other NAS methods except $\beta$-DARTS. Besides, we find that, under almost all proxies, $\beta$-DARTS can always achieve the best performance among \bp{all the} compared NAS methods. However, drastically reducing each kind of proxy still degrades the performance of $\beta$-DARTS inevitably. Considering that the learning of network weights is directly related to various proxies, we attempt to further improve the proxy robustness of $\beta$-DARTS by weight regularization (termed as Bi-level regularization). Recently,~\cite{ishida2020we} finds that learning until zero loss in deep networks is harmful and proposes a flooding regularization for solving the caused model overconfidence problem. Inspired by this, we introduce the novel flooding regularization to the NAS field and verify its effectiveness, for the first time. More interestingly, although flooding regularization is originally designed for model overconfidence, we find it can \bp{further} improve the search robustness of $\beta$-DARTS under various proxies, we \bp{also} provide comprehensive experimental verification and theoretical analysis \bp{on this aspect}.

The comprehensive comparisons of various NAS methods are concluded in Table~\ref{tab:total_comp}. Compared to our \bp{previous} CVPR oral conference version~\cite{ye2022beta}, the improved version (termed $\beta$-DARTS++) (1) further demonstrates the stability and generalization of DARTS with beta-decay regularization (e.g., Sec~\ref{sec:mobile}: MobileNet Search Space; Sec~\ref{sec:generalization}: Cross Datasets Generalization; and Sec~\ref{sec:combine}: Combination With Other Variations), (2) further explores the robustness of different NAS under \bp{different} kinds of proxies, and verifies the effectiveness of weight flooding regularization for improving the proxy robustness from both experimental and theoretical perspective (e.g., Sec~\ref{sec:related-proxies}: Proxies Are Non-negligible on NAS; Sec~\ref{sec:related-dependece}: Dependence of NAS on Training and Data; Sec~\ref{sec:method-flooding}: Weight Flooding Regularization; Sec~\ref{sec: proxy robustness}: The Robustness of Different NAS Methods under
Various Proxies; Sec~\ref{sec:experiments-more}: More Discussions And Findings).

In summary, our \bp{novel} search scheme, while simple, \bp{exhibits} the following outstanding properties:
 \begin{itemize}
\vspace{-3pt}
\item The search trajectories of $\beta$-DARTS and its $\beta$-variants on NAS-Bench-201 and NAS-Bench-1Shot1 show that, the found architecture has continuously rising performance, and the search process can reach its optimal point at an early epoch.
\vspace{3pt}
\item We only need to search once on the proxy dataset (e.g., CIFAR-10), but the searched architecture can obtain promising performance on various datasets (e.g., CIFAR-10, CIFAR-100, and ImageNet).
\vspace{3pt}
\item The search robustness under various proxies can be improved by a large margin, thus the cost of manually setting the optimal proxy for different tasks is reduced, and searching for proxy-limited tasks like medical image processing becomes achievable.
\vspace{3pt}
\item We can greatly cut down the searching cost, time, and memory consumption by setting reduced proxy data, proxy epochs, proxy layers and proxy channels while maintaining comparable performance.
\end{itemize}

The remainder of this paper is organized as follows. In section 2, we introduces related works of NAS. In section 3, we presents the proposed method. In section 4, we shows the comprehensive experiments and analysis. In section 5, we gives more discussions and findings. In section 6, we concludes this paper and points out possible future works.

\section{Related Works} \label{sec:related work} 
\subsection{Stability of DARTS}
DARTS based methods chronically suffer from the performance collapse issue, which is caused by the domination of parameter-free operators.
A lot of works are dedicated to resolving it. P-DARTS~\cite{pdarts} and DARTS+~\cite{darts+} directly limit the number of skip connections. Such handcrafted rules are somewhat suspicious and may mistakenly reject good networks. RDARTS~\cite{rdarts} finds that the Hessian eigenvalues can be regarded as an indicator for the collapse, and employs stronger regularization or augmentation on the training of supernet weights to reduce this value. Then SDARTS~\cite{sdarts} implicitly regularizes this indicator by adding perturbations to architecture parameters via random smoothing or adversarial attack. Both methods are indirect solutions and rely heavily on the quality of the indicator. FairDARTS~\cite{fairdarts} avoids operation competition by weighting each operation via independent sigmoid function, which will be pushed to zero or one by an MSE loss. DropNAS~\cite{dropnas} proposes a grouped operation dropout for the co-adaption problem and matthew effect. DOTS~\cite{dots} further uses the group operation search scheme to decouple the operation and topology search. DARTS-~\cite{darts-} factors out the optimization advantage of skip connection by adding an auxiliary one. In a word, these methods circumvent the domination effect of parameter-free operations by changing the searching and/or discretization process or adding extra parameters. Different from these works, we explore a more generic solution by explicitly regularizing the architecture parameters optimization, making original DARTS great again.

\subsection{Generalization of DARTS}
Improving the generalization ability of deep models has always been the focus of deep learning research. Recent works provide a guarantee on model generalization by minimizing loss value and loss sharpness simultaneously~\cite{foret2020sharpness}. However, the model generalization is not only related to the network weights, but also determined by its architecture. To this end, several methods attempt to improve the generalization of architectures in the field of NAS. AdaptNAS~\cite{adaptNAS} incorporates the idea of domain adaptation into the search process of DARTS, which can minimize the generalization gap of neural architectures between domains. 
MixSearch~\cite{mixsearch} uses a composited multi-domain multi-task dataset to search a generalizable architecture in a differentiable manner. On one hand, both \bp{the above} methods are built on the assumption of having multiple datasets, while our method is not built on multiple datasets. 
On the other hand, our focus is on regularizing architecture parameters, which is not investigated in AdaptNAS and MixSearch.
\subsection{Proxies Are Non-negligible on NAS} \label{sec:related-proxies}
Search proxies are ubiquitous in the NAS field, 
thus a few recent works attempt to find better proxies to achieve more efficient or effective NAS. As a pioneer, EcoNAS~\cite{zhou2020econas} systematically analyzes the effect of different proxies (e.g., epochs, data, and channels) on the rank consistency of evolutionary-based  NAS and further finds some fast and reliable proxies. PC-DARTS~\cite{pc-darts} randomly samples a subset (e.g., 1/4) of input channels for differentiable operation selection to reduce the memory cost and accelerate the search process. EPC-DARTS~\cite{xu2021partially} then introduces the side operation to improve the performance of the searched model under lower sampling rates (e.g., 1/8) of input channels. Proxy Data~\cite{na2021accelerating} finds that proxy data constructed by existing data selection methods are not always appropriate for NAS, and designs a new proxy data selection method for efficient NAS. Different from these works, we present the first attempt to investigate the robustness of different NAS under kinds of proxies, which is complementary to the above works that find better proxies for specific NAS. We further propose a simple-but-effective bi-level regularization to improve the robustness of DARTS under various proxies.

\subsection{Dependence of NAS on Training and Data}\label{sec:related-dependece}
Recently, training-free or label-free NAS has received increased attention. TE-NAS~\cite{tenas} uses the spectrum of the neural tangent kernel (NTK) and the linear region number of input space to rank architectures without any training. FreeNAS~\cite{freenas} introduces synaptic salience to evaluate and prune the network after initialization, without any labels or training. NASI~\cite{shu2021nasi} utilizes the NTK to characterize the convergence performance of candidate networks, so as to achieve fast search without training. Zero-Cost-PT~\cite{zero} proposes a training-free proxy, namely perturbation-based operation scoring, to evaluate the performance of sub-networks efficiently. In a word, these methods mainly focus on finding relevant measurements to \bp{indicate} the quality of architectures, and still have a performance gap with well-trained NAS. As an \bp{orthogonal} part, studying the dependence of existing NAS on training and data is equally important, but there are few related studies. RLNAS~\cite{rlnas} proposes a convergence-based one-shot NAS and finds that it is feasible to optimize the supernet with only random labels. Different from RLNAS, we explore the dependence of differentiable NAS on both training and data. We show that, with proper regularization, differentiable NAS can always maintain a promising search performance under a wide range of reduced (even extremely reduced)  proxy data, proxy epochs, proxy layers, and proxy channels.

\section{The Proposed Method} \label{sec:method}
\subsection{Formulation of DARTS}
Following \cite{zoph2018learning}, the original DARTS searches the structure of normal cells and reduction cells to stack the full network. Typically, a cell is defined as a directed acyclic graph (DAG) with \textit{N} nodes, where each node denotes a latent representation and the information between every two nodes is transformed by an edge. Each edge $(i, j)$ contains several candidate operations, and DARTS applies continuous relaxation via a set of learnable architecture parameters $\alpha$ to mix the outputs of different operations, converting the discrete operation selection problem into a differentiable parameter optimization problem,
\begin{equation}\label{eqn1} 
\scriptsize
  \begin{split}
    &\bar{O}^{(i, j)}(x)=\sum_{k=1}^{\left | \mathcal{O}\right |} \beta_{k}^{(i, j)} O_k(x),\quad \beta_{k}^{(i, j)}=\frac{\exp \left(\alpha_{k}^{(i, j)}\right)}{\sum_{k^{\prime}=1}^{\left | \mathcal{O}\right |} \exp \left(\alpha_{k^{\prime}}^{(i, j)}\right)}
  \end{split}
  \vspace{-6pt}
\end{equation} 
where $x$ and $\bar{O}$ are the input and mixed output of an edge, $\mathcal{O}$ is the candidate operation set, and $\beta$ denotes the softmax-activated architecture parameter set. In this way, we can perform architecture search in a differentiable manner by solving the following bi-level optimization objective,
\begin{equation} \label{eqn2}
  \begin{split}
    &\min_{\alpha}\mathcal{L}_{val}\left(w^{*}(\alpha), \alpha\right) \\
    &\text { s.t. } w^{*}(\alpha)=\arg \min _{w} \mathcal{L}_{train}(w, \alpha)
  \end{split}
  \vspace{-6pt}
\end{equation}
In practice, architecture parameters $\alpha$ and network weights $w$ are alternately updated on the validation and training datasets via gradient descent, and $w^{*}$ is approximated by one-step forward or current $w$~\cite{darts}.

\begin{figure*}[t] 
	\centering
	\includegraphics[width=1.0\linewidth]{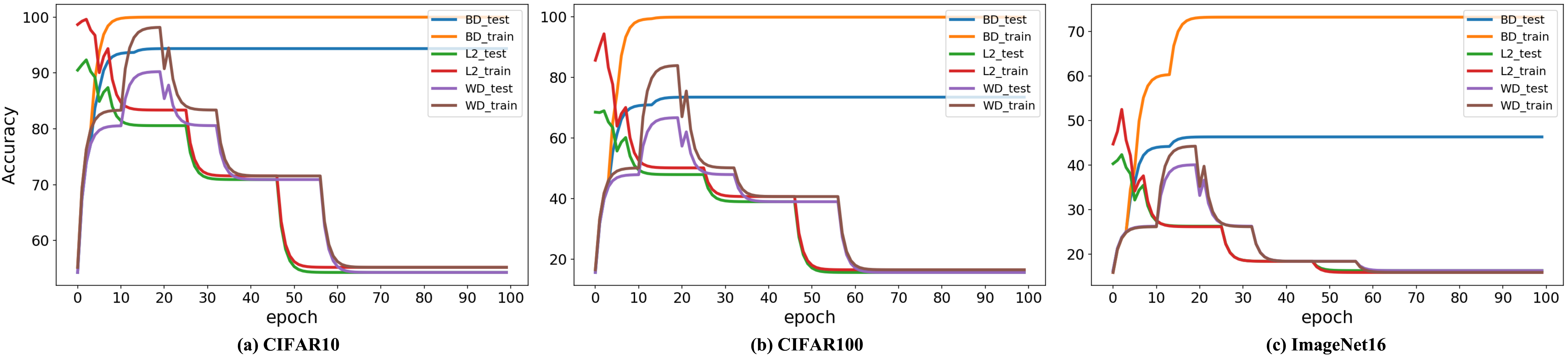}
	\vspace{-18pt}
	\caption{Accuracies of DARTS on different datasets
	 with L2, Weight Decay (WD) and Beta-Decay (BD) regularization on NAS-Bench-201 benchmark. The curve is smoothed with a coefficient of 0.5. Note that we only search once on CIFAR-10 dataset, but query the performance on different datasets. Best viewed in color.}
	\label{fig:3}
	\vspace{-14pt}	
\end{figure*}
\subsection{Beta-Decay Regularization} 
\subsubsection{Commonly-used Architecture Regularization}
In this work, we intend to improve the stability and architecture generalization of DARTS by explicitly regularizing the optimizing process of architecture parameters. Thus, we begin with the default settings of previous methods, namely L2 or weight decay regularization on architecture parameters, $\alpha$. For the convenience of analysis, we consider the single-step update of the architecture parameters,
\begin{equation} \label{eqn3}
  \begin{split}
    \alpha^{t+1}_k \leftarrow \alpha^{t}_k-\eta_{\alpha} \cdot \nabla_{\alpha_k} \mathcal{L}_{val}
  \end{split}
  \vspace{-6pt}
\end{equation}
where $\eta_{\alpha}$ and $\mathcal{L}_{val}$ are the learning rates of architecture parameters and the corresponding loss respectively. For multi-step updates, it can be transformed into a single-step update problem through step-wise recursive analysis.

In standard DARTS and its subsequent variants, Adam optimizer with L2 regularization is commonly used for architecture parameters optimization. However, for adaptive gradient algorithms, the gradients of L2 regularization are normalized ($\mathcal{N}$) by their summed magnitudes, thus the penalty for each element is relatively even, which partly offsets the effect of L2 regularization~\cite{dwd}, defined as
\begin{equation} \label{eqn4}
  \begin{split}
    \bar{\alpha}^{t+1}_k \leftarrow \alpha^{t}_k-\eta_{\alpha} \cdot \nabla_{\alpha_k} \mathcal{L}_{val}-\eta_{\alpha} \lambda \mathcal{N}\left(\alpha^{t}_k\right)
  \end{split}
  \vspace{-6pt}
\end{equation}
where $\bar{\alpha}$ denotes the single-step update with regularization. 

Considering that L2 regularization may not be effective in adaptive gradient algorithms and differs from weight decay regularization~\cite{dwd}, without loss of generality, we also include architecture parameters optimization with weight decay regularization~\cite{hanson1988comparing} for comparison, defined as
\begin{equation} \label{eqn6}
  \begin{split}
    \bar{\alpha}^{t+1}_k \leftarrow \alpha^{t}_k-\eta_{\alpha} \cdot \nabla_{\alpha_k} \mathcal{L}_{val}-\eta_{\alpha} \lambda \alpha^{t}_k
  \end{split}
  \vspace{-6pt}
\end{equation}
\subsubsection{Beta-Decay Regularization}
In~\ref{sec:alpha-beta}, we show that default L2 or weight decay regularization on $\alpha$ may not be effective or even counterproductive. Since the searching and discretization process of DARTS actually utilize softmax-activated architecture parameter set, $\beta$, to represent the importance of each operator, we shall pay more attention to the explicit regularization on $\beta$. 
As shown in Subsection~\ref{sec: beta_theoretical_analysis}, Beta regularization has the ability to improve the stability and architecture generalization of DARTS, which further denotes its significance. Although important, Beta regularization is typically ignored by previous works. This paper is devoted to filling this gap. Similar to the idea of most regularization methods, the core purpose of Beta regularization is to constrain the value of Beta from changing too much, formulated as
\begin{equation} \label{eqn8}
\small
  \begin{split}
    \bar{\beta}_{k}^{t+1}=\theta_{k}^{t+1}\left(\alpha_k^t \right)\beta_{k}^{t+1}
  \end{split}
  \vspace{-6pt}
\end{equation}

For simplicity, we use a $\theta$ function with $\alpha$ as independent variable to uniformly represent the influence of different $\alpha$ regularization on $\beta$ regularization here. To realize weight decay regularization on $\beta$ through $\alpha$, we firstly study the influence of $\alpha$ regularization on $\beta$ regularization. Recalling Eq.~(\ref{eqn4}) and Eq.~(\ref{eqn6}), we can conclude a unified formula as: 
\begin{equation} \label{eqn19}
  \begin{split}
    \bar{\alpha}_{k}^{t+1} \leftarrow \alpha_{k}^{t}-\eta_{\alpha} \nabla_{\alpha_{k}} \mathcal{L}_{val}-\eta_{\alpha} \lambda F\left(\alpha_{k}^{t}\right)
  \end{split}
  \vspace{-6pt}
\end{equation}
where $F$ denotes the effect of $\alpha$ regularization and $\lambda$ is the coefficient of regularization.

Then, we substitute Eq.~(\ref{eqn19}) and Eq.~(\ref{eqn3}) into Eq.~(\ref{eqn1}) to get $\bar{\beta}_{k}^{t+1}$ and $\beta_{k}^{t+1}$, and further divide the former by the latter to get $\theta_{k}^{t+1}\left(\alpha_k^t \right)$ of Eq.~(\ref{eqn8}). The detailed derivation can be found in \bp{Appendix A.1}.
\begin{equation} \label{eqn9}
\scriptsize
  \begin{split}
    \theta_{k}^{t+1}\left(\alpha_k^t \right)=\frac{\bar{\beta}_{k}^{t+1}}{\beta_{k}^{t+1}}=\frac{\sum_{k^{\prime}=1}^{\left | \mathcal{O}\right |} \exp \left(\alpha_{k^{\prime}}^{t+1}\right)}{\sum_{k^{\prime}=1}^{\left | \mathcal{O}\right |} \left[\exp \left(F(\alpha_{k}^{t})-F(\alpha_{k^{\prime}}^{t})\right)\right]^{\lambda \eta_{\alpha}} \exp \left(\alpha_{k^{\prime}}^{t+1}\right)}
  \end{split}
  \vspace{-6pt}
\end{equation}

As we can see in Eq.~(\ref{eqn9}), the mapping function $F$ determines the influence of $\alpha$ on $\beta$. Thus, all we need is to look for a suitable mapping function, $F$. Intuitively, a satisfactory $F$ should meet the following two \bp{conditions}: (1) $F$ is not affected by the amplitude of $\alpha$ (to avoid invalid regularization and optimization difficulties). 
(2) $F$ can reflect the relative amplitude of $\alpha$ (to impose more penalty on larger amplitude). 
To satisfy \bp{these} two requirements, we adopt the softmax to normalize $\alpha$, 
\begin{equation} \label{eqn11}
  \begin{split}
    F\left(\alpha_{k}\right)=\frac{\exp \left(\alpha_{k}\right)}{\sum_{k^{\prime}=1}^{\left | \mathcal{O}\right |} \exp \left(\alpha_{k^{\prime}}\right)}
  \end{split}
  \vspace{-6pt}
\end{equation}
Then, we introduce our proposed Beta-Decay regularization loss, whose gradients with respective to $\alpha$ equals $F\left(\alpha\right)$, 
\begin{equation} \label{eqn10}
\small
  \begin{split}
    \mathcal{L}_{Beta} = \log \left(\sum_{k=1}^{\left | \mathcal{O}\right |}  e^{\alpha_k}\right) = \operatorname{smoothmax} \left(\left\{\alpha_{k}\right\}\right) 
  \end{split}
  \vspace{-6pt}
\end{equation}

After that, substituting Eq.~(\ref{eqn11}) into Eq.~(\ref{eqn9}), we can obtain the following equation, which further accounts for the effect of Beta-Decay regularization,
\begin{equation} \label{eqn12}
\scriptsize
  \begin{split}
    \theta_{k}^{t+1}\left(\alpha_k^t \right)=\frac{\sum_{k^{\prime}=1}^{\left | \mathcal{O}\right |} \exp \left(\alpha_{k^{\prime}}^{t+1}\right)}{\sum_{k^{\prime}=1}^{\left | \mathcal{O}\right |} \left[\exp \left(\frac{\exp(\alpha_{k}^{t})-\exp(\alpha_{k^{\prime}}^{t})}{\sum_{k^{\prime\prime}=1}^{\left | \mathcal{O}\right |} \exp \left(\alpha_{k^{\prime\prime}}^{t}\right)}\right)\right]^{\lambda \eta_{\alpha}} \exp \left(\alpha_{k^{\prime}}^{t+1}\right)}
  \end{split}
  \vspace{-6pt}
\end{equation}
From the above formula, we can get the following conclusions: (1) When $\alpha$ is the largest, $\theta$ is the smallest and less than 1; when $\alpha$ is the smallest, $\theta$ is the largest and greater than 1; and when $\alpha$ is equal to itself,
$\theta=1$. (2) In current iteration, $\theta$ decreases as $\alpha$ increases. (3) $\theta$ is smaller when $\alpha$ is larger, and $\theta$ is larger when $\alpha$ is smaller. As a result, the variance of $\beta$ is constrained to be smaller, and the value of $\beta$ is constrained to be closer to its mean, achieving the similar effect with weight decay regularization of $\beta$, thus called Beta-Decay regularization.
\subsubsection{Theoretical Analysis}\label{sec: beta_theoretical_analysis}
  \begin{figure}[t] 
	\centering
	\includegraphics[width=2.8in]{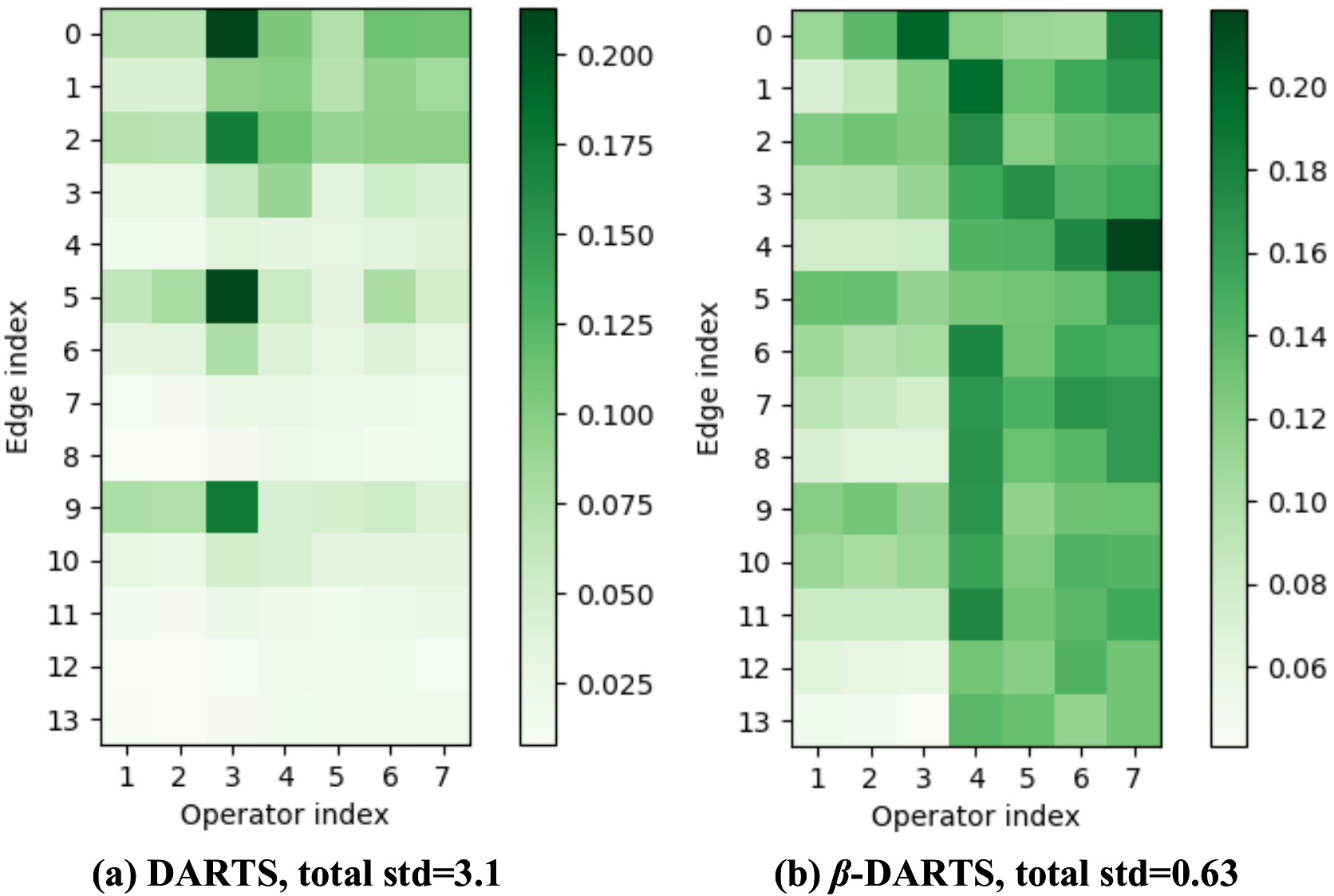}
	\vspace{-6pt}
	\caption{The beta distribution of normal cell learned by DARTS and $\beta$-DARTS, on the original search space in CIFAR-10. The operator index 1, 2, and 3 means the max pooling, avg pooling, and skip connect, while others are the parametric operators. The total std is calculated by the sum of the standard deviation of all edges under the edge independence assumption. Best viewed in color.} %
	\label{fig:4}
	\vspace{-14pt}	
\end{figure}

\begin{figure*}[t] 
	\centering
	\includegraphics[width=1.0\linewidth]{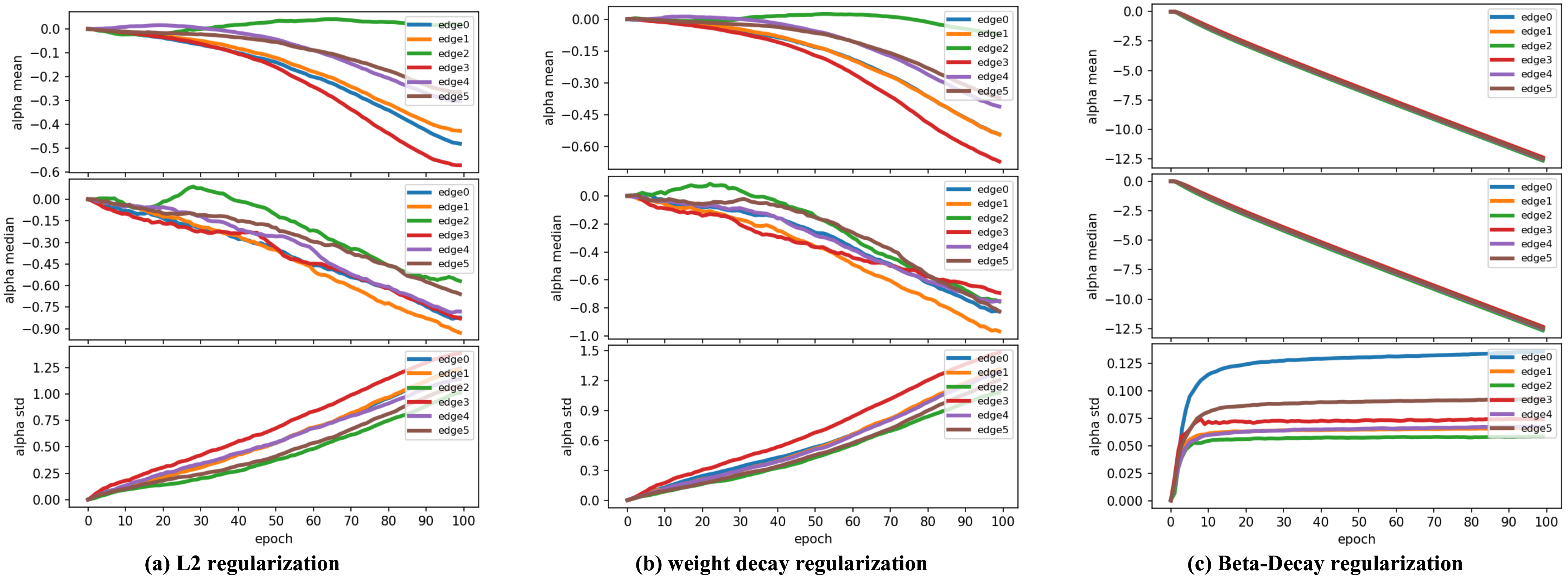}
	\vspace{-18pt}
	\caption{The alpha statistical characteristics (i.e. mean, median and standard deviation) of different edges of each epoch when searching on NAS-Bench-201 benchmark with (a) L2 regularization, (b) weight decay regularization, and (c) Beta-Decay regularization. Best viewed in color.}
	\label{fig:2}
	\vspace{-14pt}	
\end{figure*}

\noindent\textbf{Stronger Stability.} According to the theorem revealed by the recent work~\cite{prdarts}, the convergence of network weights $w$ can heavily rely on $\beta_{skip}$ in the supernet. In detail, supposing that there are three operations (convolution, skip connection and none) in the search space and the training loss is MSE, when fixing architecture parameters to optimize network weights via gradient descent, at one step the training loss can be reduced by ratio  $(1-\eta_w\varphi/4)$ with \bp{a} probability of at least $1-\sigma$, where $\eta_w$ is the corresponding learning rate and will be bounded by $\sigma$, and $\varphi$ obeys
\begin{equation} \label{eqn13}
  \begin{split}
    \varphi \propto \sum_{i=0}^{h-2}\left[\left(\beta_{conv}^{(i, h-1)}\right)^{2} \prod_{t=0}^{i-1}\left(\beta_{skip}^{(t, i)}\right)^{2}\right]
  \end{split}
  \vspace{-6pt}
\end{equation}
where $h$ is the number of supernet layers. From Eq.~(\ref{eqn13}), we can see that $\varphi$ depends more on $\beta_{skip}$ than $\beta_{conv}$, which demonstrates that the supernet weights can converge much faster with large $\beta_{skip}$. However, by imposing Beta-Decay regularization, we can redefine Eq.~(\ref{eqn13}) as follows
\begin{equation} \label{eqn14}
\small
  \begin{split}
    \varphi \propto \sum_{i=0}^{h-2}\left[\left(\theta_{conv}^{(i, h-1)}\beta_{conv}^{(i, h-1)}\right)^{2} \prod_{t=0}^{i-1}\left(\theta_{skip}^{(i, h-1)}\beta_{skip}^{(t, i)}\right)^{2}\right]
  \end{split}
  \vspace{-6pt}
\end{equation}
As mentioned before, $\theta$ becomes smaller when $\beta$ is larger and $\theta$ becomes larger when $\beta$ is smaller, which makes the convergence of network weights rely more on $\beta_{conv}$ and less on $\beta_{skip}$. \textit{From the perspective of convergence theorem}~\cite{prdarts}\textit{, the Beta-Decay regularization constrains the privilege of $\beta_{skip}$ and ensures the fair competition among architecture parameters.} As shown in Fig.~\ref{fig:3}, DARTS with L2 or weight decay regularization suffers from the performance collapse issue, while DARTS with Beta-Decay regularization has a \bp{more} stable search process. As shown in Fig.~\ref{fig:4}, the original DARTS is dominated by skip connections while $\beta$-DARTS tends to favor parametric operators.

\noindent\textbf{Stronger Generalization.} Referring to~\cite{neyshabur2017exploring} and~\cite{finlay2018lipschitz}, Lipschitz constraint is commonly used to measure and improve the generalization ability of the trained deep models. Specifically, suppose the function fitted by a deep model is $f_{w}\left(x\right)$ where $x$ is the input, when $\left\|x_{1}-x_{2}\right\|$ is very small, a well-trained model should meet the following constraint.
\begin{equation} \label{eqn15}
  \begin{split}
    \left\|f_{w}\left(x_{1}\right)-f_{w}\left(x_{2}\right)\right\| \leq C(w) \cdot\left\|x_{1}-x_{2}\right\|
  \end{split}
  \vspace{-6pt}
\end{equation}
where $C(w)$ is the Lipschitz constant. The smaller the constant is, the trained model will be less sensitive to input disturbances and have better generalization ability. 

Furthermore, we can extend this theory to differentiable architecture search. For simplicity, we consider a single-layer neural network, and multi-layer neural networks can be solved through step-wise recursive analysis. Suppose the single-layer network is mixed by the operation set $F(x)=\left(f_{1}(x),f_{2}(x),f_{3}(x)\right)$, with the corresponding architecture parameters $\beta=\left(\beta_{1}, \beta_{2}, \beta_{3}\right)$. According to Cauchy's inequality, we can get the following inequality.
\begin{equation} \label{eqn16}
\small
  \begin{split}
    &\left\|\beta F^\mathrm{T}\left(x_{1}\right)-\beta F^\mathrm{T}\left(x_{2}\right)\right\| \leq\|\beta \|\left\|F^\mathrm{T}\left(x_{1}\right)-F^\mathrm{T}\left(x_{2}\right)\right\| 
  \end{split}
  \vspace{-6pt}
\end{equation}
where $\|\beta\|=\sqrt{\sum \beta_{i}^{2}}$ can be regarded as Lipschitz constant and $\sum \beta_{i}=1$. \textit{As a result, the smaller the measure $\|\beta\|$ is, the supernet will be less sensitive to the impact of input on the operation set, and the searched architecture will have better generalization ability.} 
As shown in Fig.~\ref{fig:3}, the model searched by $\beta$-DARTS on CIFAR-10 can well generalize to the CIFAR-100 and ImageNet16 datasets and achieve excellent results. As shown in Fig.~\ref{fig:2} and Fig.~\ref{fig:4}, the architecture parameter distribution learned by $\beta$-DARTS maintains a relatively small standard deviation, ensuring the generalization ability of the searched model.

\subsubsection{Common Architecture Regularization May Not Work} \label{sec:alpha-beta}
When using L2 regularization on $\alpha$, we can obtain its effect on $\beta$ according to Eq.~(\ref{eqn4}) and Eq.~(\ref{eqn9}), defined as
\begin{equation} \label{eqn5}
\footnotesize
  \begin{split}
    \frac{\bar{\beta}_{k}^{t+1}}{\beta_{k}^{t+1}}=\frac{\sum_{k^{\prime}=1}^{\left | \mathcal{O}\right |} \exp \left(\alpha_{k^\prime}^{t+1}\right)}{\sum_{k^{\prime}=1}^{\left | \mathcal{O}\right |} \left[\exp \left(\mathcal{N}(\alpha_{k}^{t})-\mathcal{N}(\alpha_{k^{\prime}}^{t})\right)\right]^{\lambda \eta_{\alpha}} \exp \left(\alpha_{k^\prime}^{t+1}\right)} 
  \end{split}
  \vspace{-6pt}
\end{equation}

Similarly, when using weight decay on $\alpha$, we can obtain its effect on $\beta$ according to Eq.~(\ref{eqn6}) and Eq.~(\ref{eqn9}), as follows
\begin{equation} \label{eqn7}
\footnotesize
  \begin{split}
    \frac{\bar{\beta}_{k}^{t+1}}{\beta_{k}^{t+1}}=\frac{\sum_{k^{\prime}=1}^{\left | \mathcal{O}\right |} \exp \left(\alpha_{k^{\prime}}^{t+1}\right)}{\sum_{k^{\prime}=1}^{\left | \mathcal{O}\right |} \left[\exp \left(\alpha_{k}^{t}-\alpha_{k^{\prime}}^{t}\right)\right]^{\lambda \eta_{\alpha}} \exp \left(\alpha_{k^{\prime}}^{t+1}\right)}
  \end{split}
  \vspace{-6pt}
\end{equation}

From Eq.~(\ref{eqn5}) and Eq.~(\ref{eqn7}), we can find that: (1) When the values in $\alpha$ are all around 0, achieving the purpose of L2 and weight decay regularization, Alpha regularization has little effect on Beta; while when the values of $\alpha$ are all not near 0, it means that both regularizations do not work. (2) For L2 and weight decay regularization on $\alpha$, only when the median of $\alpha$ is equal to 0, Alpha regularization has the same and correct effect as Beta regularization. (3) A large variance of $\alpha$ is undesirable, which conflicts with the purpose of L2 and weight decay regularization, and makes the optimization process more sensitive to the hyperparameters $\lambda$ and $\eta_{\alpha}$. In addition, we further show the alpha statistical characteristics when searching with different regularization in Fig.~\ref{fig:2}, we can see that for L2 and weight decay regularization: (1) The mean and median of $\alpha$ continue to decrease and gradually move away from 0. (2) The standard deviation of $\alpha$ increases monotonically. These mathematical and experimental results show that, L2 or weight decay regularization commonly used in existing gradient-based NAS methods is not identical to Beta regularization, and may not be effective or even counterproductive. As a comparison, with our proposed Beta-Decay regularization: (1) The mean and median of $\alpha$ are basically equal 
\bp{since for different edges at different epochs, their changing trends tend to be similar.} (2) When the standard deviation of $\alpha$ increases to a certain extent \bp{after some epochs}, it will remain unchanged until the last epoch.

\subsection{Weight Flooding Regularization}\label{sec:method-flooding}
\subsubsection{Motivation}
On the one hand, search proxies are ubiquitous in the NAS field, like proxy search data, proxy supernet settings, and proxy training strategies. On the other hand, proxies are non-negligible as they affect the computation and memory cost as well as the search effectiveness and efficiency. Therefore, a few recent works~\cite{zhou2020econas, xu2021partially, na2021accelerating} propose to find better proxies to realize more efficient NAS. Different from these works, we propose to investigate the robustness of different NAS under kinds of proxies for the first time. As such a new perspective is commonly ignored by previous works, we first benchmark different NAS methods including our $\beta$-DARTS, as shown in Sec~\ref{sec: proxy robustness}. Then, considering network weights widely exist in the supernet of differentiable NAS, we attempt to further improve the search robustness of $\beta$-DARTS under various proxies by regularizing the optimization process of network weights.

\subsubsection{Commonly-used Weight Regularization}
We first conclude several commonly-used weight regularization methods. In standard DARTS and its numerous variants, SGD optimizer with L2 regularization is generally used for network weights optimization, defined as
\begin{equation} \label{w_l2}
  \begin{split}
    \arg \min _{w} (\mathcal{L}_{train}(w, \alpha)+\mu\|w\|_{2}^{2})
  \end{split}
  \vspace{-6pt}
\end{equation}
where $\mu$ is the loss coefficient of L2 regularization. Note that, for the SGD optimizer, weight decay regularization is theoretically equivalent to L2 regularization.

RDARTS~\cite{rdarts} finds that the dominant eigenvalue of the Hessian matrix $\nabla_{\alpha}^2 \mathcal{L}_{val}$ is related to the collapse problem. Then, employing larger regularization or stronger augmentation in the training of supernet weights is proposed to implicitly regularize this value and improve the performance of the searched model. Since the performance improvement achieved with larger regularization or stronger augmentation is basically the same, for simplicity, we only show the former (termed RDARTS-L2) as 
\begin{equation} \label{w_larger_l2}
  \begin{split}
\arg \min _{w} (\mathcal{L}_{train}(w, \alpha)+\varphi\|w\|_{2}^{2})
  \end{split}
  \vspace{-6pt}
\end{equation}
where $\varphi>>\mu$ denotes the larger regularization.  

Instead of minimizing $w$ via the current $\alpha$, the SDARTS~\cite{sdarts} proposes to minimize $w$ within a neighborhood of $\alpha$, so as to implicitly regularize the Hessian norm of the validation loss $\mathcal{L}_{val}$ and smooth the loss landscape. In practice, before each update step of network weights, SDARTS would add small perturbations to architecture parameters via random smoothing (Eq.~(\ref{w_rs})) or adversarial attack (Eq.~(\ref{w_adv})), defined as
\begin{equation} \label{w_rs}
  \begin{split}
 &\arg \min _{w} E_{\delta \sim U_{[-\epsilon, \epsilon]}} \mathcal{L}_{train}(w, \alpha+\delta) \\ 
  \end{split}
\end{equation}
\vspace{-18pt}
\begin{equation} \label{w_adv}
  \begin{split} 
 &\arg \min _{w} \max _{\|\delta\| \leq \epsilon} \mathcal{L}_{train}(w, \alpha+\delta) 
  \end{split}
  \vspace{-6pt}
\end{equation}

\subsubsection{Weight Flooding Regularization}
Since the learning of network weights is directly related to proxy data, proxy supernet settings, and proxy training strategies, we shall pay more attention to the explicit regularization on $w$ when considering the proxy robustness. However, only using common weight L2 regularization in differentiable NAS may not well handle various proxies, as will be shown in Sec~\ref{sec: proxy robustness}. Although RDARTS~\cite{rdarts} and SDARTS~\cite{sdarts} propose alternative weight regularization methods, their purposes are to regularize the validation loss $\mathcal{L}_{val}$ instead of the training loss $\mathcal{L}_{train}$. Besides, all the above methods constantly minimize the loss value of $\mathcal{L}_{train}$, providing few guarantees on weight robustness.

Recently, ~\cite{ishida2020we} finds that learning until zero loss in deep networks may cause the problem of model overconfidence and test performance degradation. To solve this problem, ~\cite{ishida2020we} proposes the flooding regularization to purposely prevent the loss from further decreasing once it reaches a reasonably small value (namely flood level). With the help of flooding regularization, the deep model is expected to reach a flat loss landscape by continuing to “random walk” around the flood level. Inspired by this, we further introduce the flooding regularization into the network weights optimization of our $\beta$-DARTS. To our best knowledge, we are the first to explore the flooding regularization in the NAS field and verify its effectiveness. More interestingly, although flooding regularization is originally designed for model overconfidence, we find it can improve the search robustness of $\beta$-DARTS under various proxies, we further give the experimental verification in Sec~\ref{sec: proxy robustness} and the theoretical analysis in the following subsection. Formally, we define the flooding regularization of supernet weights as
\begin{equation} 
  \begin{split}
\arg \min _{w} (|\mathcal{L}_{train}(w, \alpha)-b|+b)
  \end{split}
  \vspace{-6pt}
\end{equation}
where $b$ is the flood level. The training of network weights will do mini-batched gradient descent as usual but gradient ascent if $\mathcal{L}_{train}$ is below $b$. 

\subsubsection{Theoretical Analysis}
\noindent\textbf{Stronger Robustness.}   
Taking a closer inspection, when the training loss is around the flood level $t$, the training process begins to alternately perform gradient descent and gradient ascent. Suppose that the model performs gradient descent for the $n$-th training iteration and then gradient ascent for iteration $n+1$, which is defined as 
\begin{equation} \label{descent_ascent}
  \begin{split}
&w_{n}=w_{n-1}-\eta_{w} g\left(w_{n-1}\right) \\
&w_{n+1}=w_{n}+\eta_{w} g\left(w_{n}\right)
  \end{split}
  \vspace{-6pt}
\end{equation}
where $\eta_{w}$ is the learning rate of network weights, and $g\left(w\right)$ represents $\nabla_{w} \mathcal{L}_{train}(w, \alpha)$. Note that, the following analysis also applies to the case of first gradient ascent and then gradient descent. Based on Eq.~(\ref{descent_ascent}), we can get
\begin{equation} \label{descent_ascent_1}
\small
  \begin{split}
w_{n+1} & =w_{n-1}-\eta_{w} g\left(w_{n-1}\right)+\eta_{w} g\left(w_{n-1}-\eta_{w} g\left(w_{n-1}\right)\right)
  \end{split}
  \vspace{-6pt}
\end{equation}

Further, by approximating with the Tayler expansion
\begin{equation} \label{talyer_expansion}
  \begin{split} 
& g\left(w_{n-1}-\eta_{w} g\left(w_{n-1}\right)\right) \\ & \approx g\left(w_{n-1}\right)-\eta_{w} \nabla_{w} g\left(w_{n-1}\right) g\left(w_{n-1}\right)
  \end{split}
  \vspace{-6pt}
\end{equation}

Last, substituting Eq.~(\ref{talyer_expansion}) into Eq.~(\ref{descent_ascent_1}), we can obtain 
\begin{equation}
  \begin{split}
w_{n+1}&=w_{n-1}-\eta_{w}^{2} \nabla_{w}g\left(w_{n-1}\right) g\left(w_{n-1}\right) \\
&=w_{n-1}-\frac{\eta_{w}^{2}}{2} \nabla_{w}\left\|g\left(w_{n-1}\right)\right\|^{2}
  \end{split}
  \vspace{-6pt}
\end{equation}

\begin{table*}[t]
\begin{center}
\caption{Performance comparison on NAS-Bench-201 benchmark~\cite{dong2020bench}. Note that the proposed $\beta$-DARTS only searches on CIFAR-10 dataset, but can robustly achieve new SOTA on CIFAR-10, CIFAR-100 and ImageNet16-120 datasets. Averaged on 4 independent runs of searching.}
\vspace{-6pt}
\label{tab:1}
\small
\begin{tabular}{lccccccc}
\hline
\multirow{2}{*}{Methods} & \multirow{2}{*}{\begin{tabular}[c]{@{}c@{}}Cost\\ (hours)\end{tabular}} & \multicolumn{2}{c}{CIFAR-10}              & \multicolumn{2}{c}{CIFAR-100}             & \multicolumn{2}{c}{ImageNet16-120}        \\ \cline{3-8} 
                         &                                                                         & valid               & test                & valid               & test                & valid               & test                \\ \hline
DARTS(1st)~\cite{darts}               & 3.2                                                                     & 39.77±0.00          & 54.30±0.00          & 15.03±0.00          & 15.61±0.00          & 16.43±0.00          & 16.32±0.00          \\
DARTS(2nd)~\cite{darts}               & 10.2                                                                    & 39.77±0.00          & 54.30±0.00          & 15.03±0.00          & 15.61±0.00          & 16.43±0.00          & 16.32±0.00          \\
GDAS~\cite{GDAS}                     & 8.7                                                                     & 89.89±0.08          & 93.61±0.09          & 71.34±0.04          & 70.70±0.30          & 41.59±1.33          & 41.71±0.98          \\
SNAS~\cite{snas}                     & -                                                                       & 90.10±1.04          & 92.77±0.83          & 69.69±2.39          & 69.34±1.98          & 42.84±1.79          & 43.16±2.64          \\
DSNAS~\cite{dsnas}                    & -                                                                       & 89.66±0.29          & 93.08±0.13          & 30.87±16.40         & 31.01±16.38         & 40.61±0.09          & 41.07±0.09          \\
PC-DARTS~\cite{pc-darts}                 & -                                                                       & 89.96±0.15          & 93.41±0.30          & 67.12±0.39          & 67.48±0.89          & 40.83±0.08          & 41.31±0.22          \\
iDARTS~\cite{idarts}                   & -                                                                       & 89.86±0.60          & 93.58±0.32          & 70.57±0.24          & 70.83±0.48          & 40.38±0.59          & 40.89±0.68          \\
DARTS-~\cite{darts-}                   & 3.2                                                                     & 91.03±0.44          & 93.80±0.40          & 71.36±1.51          & 71.53±1.51          & 44.87±1.46          & 45.12±0.82          \\
CDARTS~\cite{CDARTS}                   & -                                                                     & 91.12±0.44          & 94.02±0.31          & 72.12±1.23          & 71.92±1.30          & 45.09±0.61          & 45.51±0.72          \\
$\beta$-DARTS               & 3.2                                                                     & \textbf{91.55±0.00} & \textbf{94.36±0.00} & \textbf{73.49±0.00} & \textbf{73.51±0.00} & \textbf{46.37±0.00} & \textbf{46.34±0.00} \\
optimal                  & -                                                                       & 91.61               & 94.37               & 73.49               & 73.51               & 46.77               & 47.31               \\ \hline
\end{tabular}
\end{center}
\vspace{-15pt}
\end{table*}

Theoretically, with the flooding regularization, the model first conducts common training as usual, then when the loss is relatively low, it changes to a new training mode where the objective is to minimize $\left\|g\left(w_{n-1}\right)\right\|^{2}$ and the learning rate is $\frac{\eta_{w}^{2}}{2}$. \textit{As a result, the network weights will be pushed to a more robust area with a smoother parameter landscape, which further makes the search robust to various proxies.} 

\section{Experiments and Analysis}
In this section, we conduct extensive experiments on various search spaces (i.e. NAS-Bench-201, DARTS, MobileNet and NAS-Bench-1Shot1) and datasets (i.e. CIFAR-10, CIFAR-100, ImageNet16-120 and ImageNet-1K). We firstly verify the stability and generalization of DARTS with Beta-Decay regularization (termed as $\beta$-DARTS) on various benchmarks. Then, we show the robustness of DARTS with Bi-Level regularization (termed as $\beta$-DARTS++) under various proxies. Last, we give ablation studies to further show the advantages of the proposed regularization methods. The overall process of $\beta$-DARTS++ is summarized in Alg~\ref{alg:Framwork}. 

\begin{algorithm}[htb]
\caption{$\beta$-DARTS++}   
\label{alg:Framwork}   
\begin{algorithmic}[1] 
\REQUIRE ~~\\ 
Architecture parameters $\alpha$; Network weights $w$; Number of search epochs $E$; Beta-Decay coefficient $\lambda_e, e\in \{1,2,…,E\}$ and Flooding coefficient $b$.
\STATE Construct a supernet and initialize architecture parameters $\alpha$ and supernet weights $w$
\label{ code:fram:extract }
\STATE For each $e\in \left[ 1, E \right]$ do   
\STATE ~~~Update architecture parameters $\alpha$ by descending \\ 
       ~~~$\nabla_\alpha (\mathcal{L}_{val}+ \lambda_e\mathcal{L}_{Beta})$
\STATE ~~~Update network weights w by descending $\nabla_{w} \mathcal{L}_{train}$ \\
~~~ or $\nabla_{w} (|\mathcal{L}_{train}-b|+b)$ ~~~ \# for reduced proxies
\STATE Derive the final architecture based on the learned $\alpha$.
\end{algorithmic} 
\end{algorithm} 

\begin{table*}[t]
\begin{center}
\caption{Comparison of SOTA models on \textbf{CIFAR-10/100(left)} and \textbf{ImageNet-1K(right)}. For CIFAR-10/100, results in the top block are obtained by training the best searched model while the bottom block shows the average results of multiple runs of searching. $^\ddagger$ denotes the results of independently searching 3 times on CIFAR-100 and evaluating on both CIFAR-10 and CIFAR-100, while $^\dagger$ denotes the results on CIFAR-10. Because of the difference in classifiers, the network parameters on CIFAR-100 are slightly more than that of CIFAR-10 (about 0.05M). For ImageNet, the top block denotes networks are directly searched on ImageNet (Img.),  and the middle block indicates architectures are searched via the idea of Cross Domain (CD.) 
using CIFAR-10 and part of ImageNet, and models in the bottom block are transferred from the searching results of CIFAR-10 (C10) or CIFAR-100 (C100). $^\ast$ denotes the model is obtained on a different search space.}
\vspace{-3pt}
\label{tab:2}
\resizebox{\linewidth}{!}{
\begin{tabular}{lcccccllccccc}
\cline{1-6} \cline{8-13}
\multirow{2}{*}{Method} & \multirow{2}{*}{\begin{tabular}[c]{@{}c@{}}GPU\\ (Days)\end{tabular}} & \multicolumn{2}{c}{CIFAR-10} & \multicolumn{2}{c}{CIFAR-100} &  & \multirow{2}{*}{Method}             & \multirow{2}{*}{\begin{tabular}[c]{@{}c@{}}GPU\\ (Days)\end{tabular}} & \multirow{2}{*}{\begin{tabular}[c]{@{}c@{}}Params\\ (M)\end{tabular}} & \multirow{2}{*}{\begin{tabular}[c]{@{}c@{}}FLOPs\\ (M)\end{tabular}} & \multirow{2}{*}{\begin{tabular}[c]{@{}c@{}}Top1\\ (\%)\end{tabular}} & \multirow{2}{*}{\begin{tabular}[c]{@{}c@{}}Top5\\ (\%)\end{tabular}} \\ \cline{3-6}
                        &                                                                       & Params(M)    & Acc(\%)       & Params(M)     & Acc(\%)       &  &                                     &                                                                       &                                                                       &                                                                      &                                                                      &                                                                      \\ \cline{1-6} \cline{8-13} 
NASNet-A~\cite{nasnet}                & 2000                                                                  & 3.3          & 97.35         & 3.3           & 83.18         &  & MnasNet-92$^\ast$(Img.)~\cite{mnasnet}                    & 1667                                                                  & 4.4                                                                   & 388                                                                  & 74.8                                                                 & 92.0                                                                 \\
DARTS(1st)~\cite{darts}              & 0.4                                                                   & 3.4          & 97.00±0.14    & 3.4           & 82.46         &  & FairDARTS$^\ast$(Img.)~\cite{fairdarts}                     & 3                                                                     & 4.3                                                                   & 440                                                                  & 75.6                                                                 & 92.6                                                                 \\
DARTS(2nd)~\cite{darts}              & 1                                                                     & 3.3          & 97.24±0.09    & \textbf{-}    & \textbf{-}    &  & PC-DARTS(Img.)~\cite{pc-darts}                      & 3.8                                                                   & 5.3                                                                   & 597                                                                  & 75.8                                                                 & 92.7                                                                 \\
SNAS~\cite{snas}                    & 1.5                                                                   & 2.8          & 97.15±0.02    & 2.8           & 82.45         &  & DOTS(Img.)~\cite{dots}                          & 1.3                                                                   & 5.3                                                                   & 596                                                                  & 76.0                                                                 & 92.8                                                                 \\
GDAS~\cite{GDAS}                    & 0.2                                                                   & 3.4          & 97.07         & 3.4           & 81.62         &  & DARTS-$^\ast$(Img.)~\cite{darts-}                        & 4.5                                                                   & 4.9                                                                   & 467                                                                  & 76.2                                                                 & 93.0                                                                 \\ \cline{8-13} 
P-DARTS~\cite{pdarts}                 & 0.3                                                                   & 3.4          & 97.50         & 3.6           & 82.51         &  & \multicolumn{1}{c}{AdaptNAS-S(CD.)~\cite{adaptNAS}} & 1.8                                                                   & 5.0                                                                   & 552                                                                  & 74.7                                                                 & 92.2                                                                 \\
PC-DARTS~\cite{pc-darts}                & 0.1                                                                   & 3.6          & 97.43±0.07    & 3.6           & 83.10         &  & \multicolumn{1}{c}{AdaptNAS-C(CD.)~\cite{adaptNAS}} & 2.0                                                                   & 5.3                                                                   & 583                                                                  & 75.8                                                                 & 92.6                                                                 \\ \cline{1-6} \cline{8-13} 
P-DARTS~\cite{pdarts}                 & 0.3                                                                   & 3.3±0.21     & 97.19±0.14    & -             & -             &  & AmoebaNet-C(C10)~\cite{amoebanet}                    & 3150                                                                  & 6.4                                                                   & 570                                                                  & 75.7                                                                 & 92.4                                                                 \\
R-DARTS(L2)~\cite{rdarts}             & 1.6                                                                   & -            & 97.05±0.21    & -             & 81.99±0.26    &  & SNAS(C10)~\cite{snas}                           & 1.5                                                                   & 4.3                                                                   & 522                                                                  & 72.7                                                                 & 90.8                                                                 \\
SDARTS-ADV~\cite{sdarts}              & 1.3                                                                   & 3.3          & 97.39±0.02    & -             & -             &  & P-DARTS(C100)~\cite{pdarts}                       & 0.3                                                                   & 5.1                                                                   & 577                                                                  & 75.3                                                                 & 92.5                                                                 \\
DOTS~\cite{dots}                    & 0.3                                                                   & 3.5          & 97.51±0.06    & 4.1           & 83.52±0.13    &  & SDARTS-ADV(C10)~\cite{sdarts}                     & 1.3                                                                   & 5.4                                                                   & 594                                                                  & 74.8                                                                 & 92.2                                                                 \\
DARTS+PT~\cite{darts+pt}                & 0.8                                                                   & 3.0          & 97.39±0.08    & -             & -             &  & DOTS(C10)~\cite{dots}                           & 0.3                                                                   & 5.2                                                                   & 581                                                                  & 75.7                                                                 & 92.6                                                                 \\
DARTS-~\cite{darts-}                  & 0.4                                                                   & 3.5±0.13     & 97.41±0.08    & 3.4           & 82.49±0.25    &  & DARTS+PT(C10)~\cite{darts+pt}                       & 0.8                                                                   & 4.6                                                                   & -                                                                    & 74.5                                                                 & 92.0                                                                 \\
$\beta$-DARTS$^\ddagger$         & 0.4                                                                   & 3.78±0.08    & 97.49±0.07    & 3.83±0.08     & 83.48±0.03    &  & $\beta$-DARTS(C100)                     & 0.4                                                  & 5.4                                                  & 597                                                 & 75.8                                                 & 92.9                                                 \\
$\beta$-DARTS$^\dagger$        & 0.4                                                                    & 3.75±0.15    & 97.47±0.08    & 3.80±0.15     & 83.76±0.22    &  & $\beta$-DARTS(C10)                    & 0.4                                                   & 5.5                                                   & 609                                                   & 76.1                                                                     & 93.0                                                                     \\ \cline{1-6} \cline{8-13} 
\end{tabular}
}
\end{center}
\vspace{-12pt}
\end{table*}

\vspace{-8pt}
\subsection{The Effectiveness of Beta-Decay Regularization on Various Benchmarks}
\subsubsection{Results on NAS-Bench-201 Search Space} 
\noindent\textbf{Settings.} NAS-Bench-201~\cite{dong2020bench} is the most widely used NAS benchmark analyzing various NAS methods. NAS-Bench-201 provides a DARTS-like search space, containing 4 internal nodes with 5 associated operations. The search space consists of 15,625 architectures, with the ground truth performance of CIFAR-10, CIFAR-100 and ImageNet16-120 of each architecture provided. On NAS-Bench-201, the searching settings are kept the same as DARTS on~\cite{dong2020bench}.

\begin{table}[t]
\caption{Performance comparison on MobileNet search space. All these methods are directly searched and evaluated on \textbf{ImageNet-1K}. The bottom denotes models that have SE and Swish modules. $\star$: using the DARTS search space.}\label{tab:mobile}
\vspace{-3pt}
\centering
\setlength{\tabcolsep}{1.1mm}{
\begin{tabular}{lcccccc}
\hline
Models & \begin{tabular}[c]{@{}c@{}}FLOPs\\ (M)\end{tabular} & \begin{tabular}[c]{@{}c@{}}Params\\ (M)\end{tabular} & \begin{tabular}[c]{@{}c@{}}Top1\\ (\%)\end{tabular} & \begin{tabular}[c]{@{}c@{}}Top5\\ (\%)\end{tabular} & \begin{tabular}[c]{@{}c@{}}GPU\\ (days)\end{tabular} & \begin{tabular}[c]{@{}c@{}}Proxy\\ data\end{tabular}  \\ \hline
MnasNet-92~\cite{mnasnet}              & 388                                                                  & 3.9                                                                   & 74.8                                                                 & 92.1                                                                  & 3791                                                                   & 100\%                                                                 \\
FBNet-C~\cite{wu2019fbnet}                 & 375                                                                  & 5.5                                                                   & 74.9                                                                  & 92.3                                                                  & 9                                                                      & 100\%                                                                 \\
FairNAS-A~\cite{chu2021fairnas}               & 388                                                                  & 4.6                                                                   & 75.3                                                                  & 92.4                                                                  & 12                                                                     & 100\%                                                                 \\
SCARLET-A~\cite{chu2021scarlet}               & 365                                                                  & 6.7                                                                   & 76.9                                                                  & 93.4                                                                  & 10                                                                     & 100\%                                                                 \\
ProxylessNAS~\cite{cai2018proxylessnas}            & 465                                                                  & 7.1                                                                   & 75.1                                                                  & 92.5                                                                  & 15                                                                     & 100\%                                                                 \\
FairDARTS-D~\cite{fairdarts}             & 440                                                                  & 4.3                                                                   & 75.6                                                                  & 92.6                                                                  & 3                                                                      & 100\%                                                                 \\
DARTS-~\cite{darts-}                  & 467                                                                  & 4.9                                                                   & 76.2                                                                  & 93.0                                                                    & 4.5                                                                    & 100\%                                                                 \\
PC-DARTS$^\star$~\cite{pc-darts}            & 597                                                                  & 5.3                                                                   & 75.8                                                                  & 92.7                                                                  & 3.8                                                                    & 12.5\%                                                               \\
ER-PC-DARTS$^\star$~\cite{xu2021partially}           & 578                                                                  & 5.1                                                                   & 75.9                                                                  & 92.8                                                                  & 2.8                                                                    & 12.5\%                                                               \\ \hline
MobileNetV3~\cite{koonce2021mobilenetv3}             & 219                                                                  & 5.4                                                                   & 75.2                                                                  & 92.2                                                                  & 3000                                                                   & 100\%                                                                 \\
MoGA-A~\cite{chu2020moga}                  & 304                                                                  & 5.1                                                                   & 75.9                                                                  & 92.8                                                                  & 12                                                                     & 100\%                                                                 \\
OFA~\cite{cai2019once}                    & 230                                                                  & --                                                                    & 76.9                                                                  & --                                                                    & 55                                                                     & 100\%                                                                 \\
DNA-b~\cite{li2020block}                   & 406                                                                  & 4.9                                                                   & 77.5                                                                  & 93.3                                                                  & 24.6                                                                   & 100\%                                                                 \\
MixNet-M~\cite{tan2019mixconv}                & 360                                                                  & 5.0                                                                     & 77.0                                                                    & 93.3                                                                  & 3000                                                                   & 100\%                                                                    \\
EfficientNet B0~\cite{tan2019efficientnet}         & 390                                                                  & 5.3                                                                   & 76.3                                                                  & 93.2                                                                  & 3000                                                                   & 100\%                                                                 \\
NoisyDARTS-A~\cite{chu2020noisy}            & 449                                                                  & 5.5                                                                   & 77.9                                                                  & 94.0                                                                    & 12                                                                     & 100\%                                                                 \\
DARTS-~\cite{darts-}                  & 470                                                                  & 5.5                                                                   & 77.8                                                                  & 93.9                                                                  & 4.5                                                                    & 100\%                                                                 \\
$\beta$-DARTS                 & 476                                                                  & 4.8                                                                   & 78.1                                                                  & 93.8                                                                  & 0.8                                                                    & 12.5\%                                                               \\ \hline
\end{tabular}}
\vspace{-8pt}
\end{table}

\noindent\textbf{Results.} The comparison results are shown in Table~\ref{tab:1}. We only search on CIFAR-10 and use the found genotype to query the performance on various datasets. For stability, our 4 runs of searching under different random seeds always find the same optimal solution, which is very close to the optimal performance of NAS-Bench-201. Moreover, as shown in Fig.~\ref{fig:3}, the performance collapse issue is well solved and $\beta$-DARTS has a stable and continuously rising search process. For generalization ability, we can see that the model found on CIFAR-10 achieves consistent new SOTA on CIFAR-10, CIFAR-100 and ImageNet. For dependency on training, as shown in Fig.~\ref{fig:3}, the search process reaches its optimal point at an early stage (i.e., before 20 epochs), on different datasets. Such results validate that $\beta$-DARTS has the ability to find the optimal architecture rapidly. More interestingly, we find that the search process of different datasets reaches the optimal point in different epochs, although they belong to the same run of searching on CIFAR-10. More similar results can be found in \bp{Appendix B.1}.

\subsubsection{Results on DARTS Search Space} 
\noindent\textbf{Settings.} Common DARTS search space~\cite{darts} is also popular for evaluating various NAS methods. The search space consists of normal cells and reduction cells. Each cell has 4 intermediate nodes with 14 edges, and each edge is associated with 8 candidate operations. On the DARTS search space, all the search settings are kept the same as the original DARTS since our method only introduces simple regularization. For evaluation settings, the evaluation on \textbf{CIFAR-10/100} follows DARTS~\cite{darts} and the evaluation on \textbf{ImageNet-1K} follows P-DARTS~\cite{pdarts} and PC-DARTS~\cite{pc-darts}. 


\noindent\textbf{Results.} The comparison results are shown in Table~\ref{tab:2}. We search on CIFAR-10 or CIFAR-100 while evaluating the inferred architecture on CIFAR-10, CIFAR-100 and ImageNet. For stability, the average results of multiple independent runs of $\beta$-DARTS achieve the SOTA performance on both CIFAR-10 and CIFAR-100, namely 97.47±0.08\% and 83.48±0.03\%, without extra changes or any cost. For generalization ability, architectures found on CIFAR-100 can still yield a SOTA result of 97.49±0.07\% on CIFAR-10, and models found on CIFAR-10 obtain a new SOTA of 83.76±0.22\%  on CIFAR-100, and networks found on both CIFAR-10 and CIFAR-100 datasets can achieve comparable results on ImageNet with those of directly searching on ImageNet or using cross-domain method. All the searched architectures are visualized in \bp{Appendix C.1}. 


\subsubsection{Results on MobileNet Search Space} \label{sec:mobile}
\noindent\textbf{Settings.} MobileNet search space is a chain-structured search space and is also widely used for evaluating NAS methods. The search space contains 21 searchable layers, and each of each is composed of inverted bottleneck MBConv~\cite{sandler2018mobilenetv2} with expansion rates of {3,6} and kernel sizes of {3,5,7} and an additional skip connection to enable depth search of models. For searching settings, we keep it almost the same as~\cite{cai2018proxylessnas}, except that we add the Beta-Decay regularization and randomly sample 10\% and 2.5\% images from the training set of ImageNet for training network weights and architecture parameters respectively, and search for 30 epochs with a batch size of 1280. For evaluation settings, we follow~\cite{darts-, chu2020noisy, li2020block} using SE and Swish modules for a fair comparison with EfficientNet~\cite{tan2019efficientnet}.

\noindent\textbf{Results.} The comparison results are shown in Table~\ref{tab:mobile}. As we can see, the architecture searched by $\beta$-DARTS can achieve a Top1 accuracy of 78.1\% with only 4.8M parameters, better than all compared methods, which verifies its effectiveness on chain-structured search space. By contrast, due to the performance collapse (i.e., skip connection domination) problem, the original DARTS can only obtain a Top1 accuracy of 66.4\%~\cite{chu2020noisy,darts-}, which further verifies the strong stability of $\beta$-DARTS to this problem. Besides, we only search on 12.5\% data of the ImageNet training set and evaluate the searched architecture on the whole ImageNet, which further verifies the generalization ability of the searched model. On the other hand, since we use fewer data and epochs, we can complete the search on ImageNet in only 0.8 GPU days. Searched model is visualized in \bp{Appendix C.2}.

\begin{figure*}[t] 
    \vspace{-3pt}	
	\centering
	\includegraphics[width=0.97\linewidth]{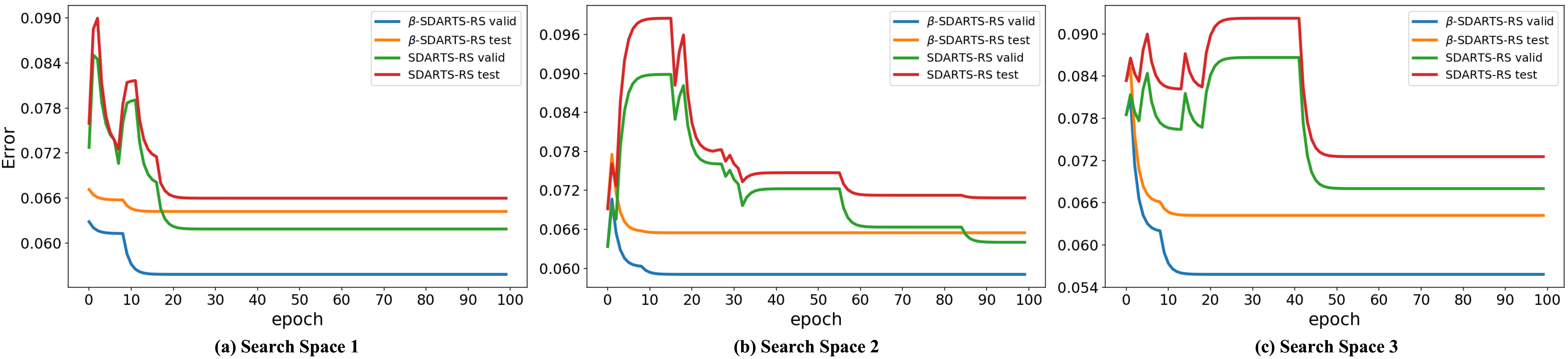}
	\vspace{-8pt}
	\caption{Test and valid error of SDARTS-RS and $\beta$-SDARTS-RS on 3 different search spaces of NAS-Bench-1Shot1~\cite{1shot1}. The curve is smoothed with a coefficient of 0.5. Best viewed in color.}
	\label{fig:6}
	\vspace{-6pt}	
\end{figure*}

\begin{table*}[t]
\begin{center}
\caption{Cross datasets verification. We show the generalization ability from C10/C100/Img16 and partial data of Img16 to CIFAR-10, CIFAR-100, and ImageNet16-120 datasets. Averaged on 3 independent runs of searching on NAS-Bench-201.}
\label{tab:cross}
\begin{tabular}{lccccccc}
\hline
\multirow{2}{*}{Methods} & \multirow{2}{*}{\begin{tabular}[c]{@{}c@{}}Data \\ (\%) \end{tabular}} & \multicolumn{2}{c}{CIFAR-10}              & \multicolumn{2}{c}{CIFAR-100}             & \multicolumn{2}{c}{ImageNet16-120}        \\ \cline{3-8} 
                         &                                                                         & valid               & test                & valid               & test                & valid               & test                \\ \hline
$\beta$-DARTS(C10)  & 100\%        & 91.55±0.00 & 94.36±0.00 & 73.49±0.00 & 73.51±0.00 & 46.37±0.00 & 46.34±0.00 \\
$\beta$-DARTS(C100)  & 100\%      & 91.35±0.18     & 93.99±0.32     & 72.32±1.02     & 72.33±1.03     & 46.17±0.37     & 45.89±1.08     \\
$\beta$-DARTS(Img16) & 100\%      & 91.55±0.00     & 94.36±0.00     & 73.49±0.00     & 73.51±0.00     & 46.37±0.00     & 46.34±0.00     \\
$\beta$-DARTS(Img16) & 75\%       & 91.55±0.00     & 94.36±0.00     & 73.49±0.00     & 73.51±0.00     & 46.37±0.00     & 46.34±0.00     \\
$\beta$-DARTS(Img16) & 50\%       & 91.55±0.00     & 94.36±0.00     & 73.49±0.00     & 73.51±0.00     & 46.37±0.00     & 46.34±0.00     \\
$\beta$-DARTS(Img16) & 25\%       & 91.55±0.00     & 94.36±0.00     & 73.49±0.00     & 73.51±0.00     & 46.37±0.00     & 46.34±0.00     \\
\textbf{optimal}     & \textbf{-} & \textbf{91.61} & \textbf{94.37} & \textbf{74.49} & \textbf{73.51} & \textbf{46.77} & \textbf{47.31} \\ \hline
\end{tabular}
\end{center}
\vspace{-15pt}
\end{table*}

\subsubsection{Results on NAS-Bench-1Shot1 Search Space} 
\noindent\textbf{Settings.} Moreover, we also utilize NAS-Bench-1Shot1~\cite{1shot1} benchmark and SDARTS-RS~\cite{sdarts} baseline to demonstrate the effectiveness of Beta-Decay regularization. NAS-Bench-1Shot1 contains 3 different search spaces, which consist of 6,240, 29,160 and 363,648 architectures with the CIFAR-10 performance provided separately. On NAS-Bench-1Shot1, both the operator of each edge and the topology of the cell need to be determined, and we keep the searching settings the same as SDARTS-RS~\cite{sdarts}.

\noindent\textbf{Results.} We show the search trajectory in Fig.~\ref{fig:6}. On one hand, $\beta$-SDARTS-RS can yield much lower test/validation error than SDARTS-RS across different search spaces. On the other hand, the error of $\beta$-SDARTS-RS keeps decreasing while the error of SDARTS-RS increases first and then decreases, validating the more stable search process of $\beta$-SDARTS-RS. Besides, the search process of $\beta$-SDARTS-RS also reaches its optimal point at an early stage (i.e., around 10 epochs), on different search spaces.

\subsubsection{Cross Datasets Generalization} \label{sec:generalization}
To demonstrate the cross datasets generalization ability of the searched architecture by $\beta$-DARTS, we conduct more experiments and analyses as follows. Actually, on the DARTS search space, the results in Table~\ref{tab:2} have shown the generalization from CIFAR-10/CIFAR-100 to CIFAR-100/CIFAR-10 and ImageNet-1K datasets. On MobileNet search space, the search results in Table~\ref{tab:mobile} have shown the generalization from partial ImageNet-1K to the whole ImageNet-1K. On NAS-Bench-201 benchmark, besides the generalization from CIFAR-10 (in Table~\ref{tab:1}), we further show the generalization from CIFAR-100 and ImageNet16-120 to CIFAR-10/CIFAR-100/ImageNet16-120 datasets in Table~\ref{tab:cross}. As we can see, the architecture found on any one dataset can well generalize to all three datasets. More interesting, we find that the model found on CIFAR-10 and ImageNet16-120 can always perform better than that found on CIFAR-100 on all three datasets, which verifies that CIFAR-10 and ImageNet16-120 are more suitable proxy datasets than CIFAR-100 (note that similar findings are also concluded in~\cite{panda2021nastransfer}). Furthermore, we use partial data (e.g. 75\%, 50\% and 25\% randomly sampled data) of ImageNet16-120 dataset for searching on NAS-Bench-201 benchmark and show the generalization to CIFAR-10/CIFAR-100/ImageNet16-120. As we can see, even using partial data, $\beta$-DARTS can still maintain a SOTA performance across different datasets, such results further verify the outstanding property of our search scheme, namely being less dependent on training data while presenting superior search generalization. 
Besides these experimental evidences, theoretical analysis in Section~\ref{sec: beta_theoretical_analysis} also shows our regularization can make the searching less sensitive to the impact of input, which makes $\beta$-DARTS embrace the great attribute of cross datasets generalization.

\begin{figure*}
\centering
\begin{minipage}[b]{0.99\linewidth}
\includegraphics[width=1\linewidth]{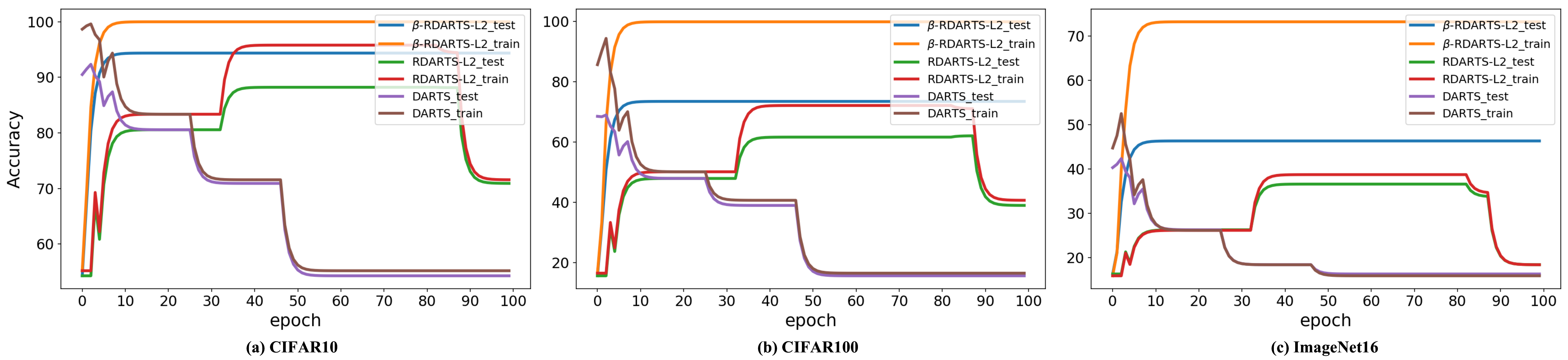}\vspace{-10pt}
\includegraphics[width=1\linewidth]{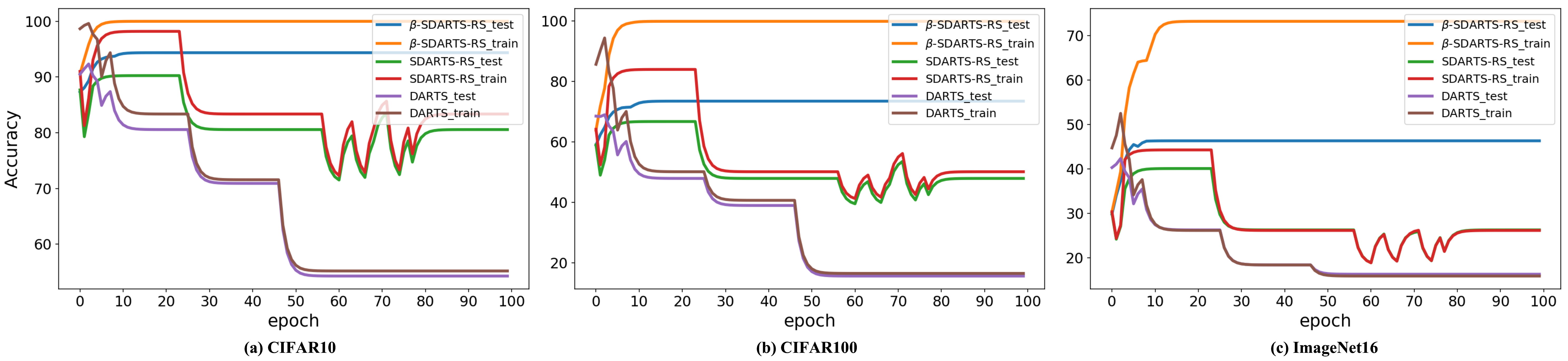}\vspace{-10pt}
\includegraphics[width=1\linewidth]{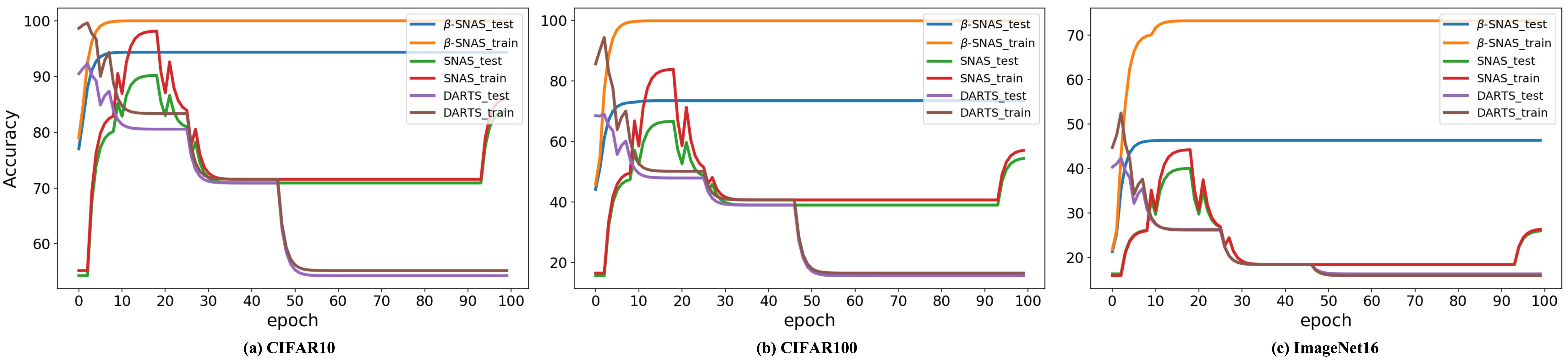}\vspace{-8pt}
\caption{Accuracies on different datasets by DARTS,  RDARTS-L2~\cite{snas} and $\beta$-RDARTS-L2, SDARTS-RS~\cite{snas} and $\beta$-SDARTS-RS, SNAS~\cite{snas} and $\beta$-SNAS on NAS-Bench-201 benchmark. The curve is smoothed with a coefficient of 0.5. Note that we only search once on CIFAR-10 dataset and report the results of different datasets.}\label{fig:beta_variants}
\end{minipage}
\vspace{-3pt}
\end{figure*}

\begin{table*}[t]
\begin{center}
\caption{The results of different DARTS variants and their Beta Decay regularization improved versions on NAS-Bench-201 benchmark. We only search on CIFAR-10 dataset and perform 3 runs of searching under different random seeds.}
\label{tab:combine}
\resizebox{\linewidth}{!}{
\begin{tabular}{lcccccc}
\hline
\multirow{2}{*}{Methods} & \multicolumn{2}{c}{CIFAR-10}                            & \multicolumn{2}{c}{CIFAR-100}                           & \multicolumn{2}{c}{ImageNet16-120}                      \\ \cline{2-7} 
                         & valid                      & test                       & valid                      & test                       & valid                      & test                       \\ \hline
SDARTS-RS~\cite{sdarts}                & 75.21/68.29/75.21          & 80.57/70.92/80.57          & 47.51/38.57/47.51          & 47.93/38.97/47.93          & 27.79/18.87/27.79          & 26.29/18.41/26.29          \\
\textbf{$\beta$-SDARTS-RS}     & \textbf{91.55/91.61/91.61} & \textbf{94.36/94.37/94.37} & \textbf{73.49/72.75/72.75} & \textbf{73.51/73.22/73.22} & \textbf{46.37/45.56/45.56} & \textbf{46.34/46.71/46.71} \\
RDARTS-L2~\cite{rdarts}               & 68.29/68.29/39.77          & 70.92/70.92/54.30          & 38.57/38.57/15.03          & 38.97/38.97/15.61          & 18.87/18.87/16.43          & 18.41/18.41/16.32          \\
\textbf{$\beta$-RDARTS-L2}     & \textbf{91.55/91.28/91.61} & \textbf{94.36/93.79/94.37} & \textbf{73.49/71.88/72.75} & \textbf{73.51/71.60/73.22} & \textbf{46.37/46.40/45.56} & \textbf{46.34/46.67/46.71} \\
SNAS~\cite{snas}                     & 82.39/89.07/91.14          & 84.16/91.89/93.60          & 54.57/67.11/71.38          & 54.64/66.99/70.74          & 27.17/39.98/44.10          & 26.10/39.13/45.03          \\
\textbf{$\beta$-SNAS}          & \textbf{91.55/91.55/91.55} & \textbf{94.36/94.36/94.36} & \textbf{73.49/73.49/73.49} & \textbf{73.51/73.51/73.51} & \textbf{46.37/46.37/46.37} & \textbf{46.34/46.34/46.34} \\
DARTS(1st)               & 39.77                      & 54.30                      & 15.03                      & 15.61                      & 16.43                      & 16.32                      \\
\textbf{optimal}         & \textbf{91.61}             & \textbf{94.37}             & \textbf{74.49}             & \textbf{73.51}             & \textbf{46.77}             & \textbf{47.31}             \\ \hline
\end{tabular}
}
\end{center}
\vspace{-12pt}
\end{table*}

\subsubsection{Combination With Other Variations} \label{sec:combine}
The proposed Beta Decay regularization can easily combine with other DARTS variants to improve both the stability of the search process and the generalization ability of the searched architecture. To this end, we also test the proposed method on SDARTS-RS~\cite{sdarts}, RDARTS-L2~\cite{rdarts} and SNAS~\cite{snas} on NAS-Bench-201 benchmark (shown as $\beta$-SDARTS-RS, $\beta$-RDARTS-L2 and $\beta$-SNAS). The implementation is consistent with their papers or open-source code. When adapting Beta Decay regularization, we set the weighting schemes of SDARTS-RS, RDARTS-L2 and SNAS as 0-50, 0-50, and 0-5 respectively.

As shown in Fig.~\ref{fig:beta_variants}, we first compare the anytime search trajectories between different baselines and their Beta Decay regularization improved versions. The search trajectory of the original DARTS is also shown in each figure. As we can see, although all these variants can improve the final result of the original DARTS, they still suffer from performance collapse/degradation issue. As a comparison, with the help of Beta Decay regularization, all these methods have continuously rising search performance and reach a much higher accuracy
. Moreover, the search processes of all the improved versions always reach the optimal point at an early stage across different datasets. In addition, we can see that the architectures searched by all these variants have a poor generalization ability (i.e., CIFAR-10 searched model performs worse on CIFAR-100 and ImageNet16-120, and its performance on CIFAR-100 and ImageNet16-120 becomes closer to the baseline), while Beta Decay regularization has the ability to relieve this problem.

As shown in Table.~\ref{tab:combine}, we further present the quantitative results between different baselines and their Beta Decay regularization improved versions. The search results of the original DARTS and the optimal results of NAS-Bench-201 are also shown in the table. As we can see, Beta Decay regularization can  significantly and stably boost the performance of all these variants by a large margin, which is close to the optimal results across various datasets. For example, SDARTS-RS and RDARTS-L2 can be improved by more than 20\% in terms of accuracy on ImageNet16-120 test set. More importantly, with Beta Decay regularization, multiple runs of searching have relatively low variance and always find the satisfactory solution, and the architecture searched on CIFAR-10 can also perform well on CIFAR-10,  CIFAR-100 and ImageNet16-120, demonstrating the stability and generalization of the proposed method.

\begin{table*}
\caption{Performance comparison under different proxy data. We benchmark different methods including DARTS~\cite{darts}, RDARTS-L2~\cite{rdarts} and SDARTS-RS~\cite{sdarts}, and compare with our methods including DARTS-flooding, $\beta$-DARTS~\cite{ye2022beta} and $\beta$-DARTS++. All the results are averaged on 3 independent runs of searching.}
\label{table:proxy-data}
    \centering
	\begin{minipage}{0.48\linewidth}
		\centering
        \subcaption{DARTS~\cite{darts}}
\resizebox{\textwidth}{18mm}{
\begin{tabular}{ccccccc}
\hline
Partial data & C10 val    & C10 test   & C100 val   & C100 test  & Img val    & Img test   \\ \hline
100\%        & 39.77±0.00 & 54.30±0.00 & 15.03±0.00 & 15.61±0.00 & 16.43±0.00 & 16.32±0.00 \\
75\%         & 39.77±0.00 & 54.30±0.00 & 15.03±0.00 & 15.61±0.00 & 16.43±0.00 & 16.32±0.00 \\
50\%         & 39.77±0.00 & 54.30±0.00 & 15.03±0.00 & 15.61±0.00 & 16.43±0.00 & 16.32±0.00 \\
25\%         & 39.77±0.00 & 54.30±0.00 & 15.03±0.00 & 15.61±0.00 & 16.43±0.00 & 16.32±0.00 \\
20\%         & 39.77±0.00 & 54.30±0.00 & 15.03±0.00 & 15.61±0.00 & 16.43±0.00 & 16.32±0.00 \\
10\%         & 39.77±0.00 & 54.30±0.00 & 15.03±0.00 & 15.61±0.00 & 16.43±0.00 & 16.32±0.00 \\
5\%          & 39.77±0.00 & 54.30±0.00 & 15.03±0.00 & 15.61±0.00 & 16.43±0.00 & 16.32±0.00 \\
2\%          & 39.77±0.00 & 54.30±0.00 & 15.03±0.00 & 15.61±0.00 & 16.43±0.00 & 16.32±0.00 \\
1\%          & 39.77±0.00 & 54.30±0.00 & 15.03±0.00 & 15.61±0.00 & 16.43±0.00 & 16.32±0.00 \\
optimal      & 91.61      & 94.37      & 73.49      & 73.51      & 46.77      & 47.31      \\ \hline
\end{tabular}}
	\end{minipage}
	\hfill
	\begin{minipage}{0.48\linewidth}
		\centering
        \subcaption{RDARTS-L2~\cite{rdarts}}
\resizebox{\textwidth}{18mm}{
\begin{tabular}{ccccccc}
\hline
Partial data & C10 val     & C10 test    & C100 val    & C100 test   & Img val    & Img test   \\ \hline
100\%        & 58.78±16.47 & 65.38±9.60  & 30.72±13.59 & 31.18±13.49 & 18.06±1.41 & 17.71±1.21 \\
75\%         & 75.45±7.29  & 78.56±6.86  & 47.07±8.28  & 47.30±8.03  & 24.51±4.91 & 23.66±4.54 \\
50\%         & 51.61±20.50 & 63.14±15.32 & 26.07±19.12 & 26.47±18.80 & 20.15±6.45 & 19.59±5.66 \\
25\%         & 39.77±0.00  & 54.30±0.00  & 15.03±0.00  & 15.61±0.00  & 16.43±0.00 & 16.32±0.00 \\
20\%         & 39.77±0.00  & 54.30±0.00  & 15.03±0.00  & 15.61±0.00  & 16.43±0.00 & 16.32±0.00 \\
10\%         & 39.77±0.00  & 54.30±0.00  & 15.03±0.00  & 15.61±0.00  & 16.43±0.00 & 16.32±0.00 \\
5\%          & 39.77±0.00  & 54.30±0.00  & 15.03±0.00  & 15.61±0.00  & 16.43±0.00 & 16.32±0.00 \\
2\%          & 39.77±0.00  & 54.30±0.00  & 15.03±0.00  & 15.61±0.00  & 16.43±0.00 & 16.32±0.00 \\
1\%          & 39.77±0.00  & 54.30±0.00  & 15.03±0.00  & 15.61±0.00  & 16.43±0.00 & 16.32±0.00 \\
optimal      & 91.61       & 94.37       & 73.49       & 73.51       & 46.77      & 47.31      \\ \hline
\end{tabular}}
	\end{minipage}
 	\qquad
  	\begin{minipage}{0.48\linewidth}
	\centering
        \vspace{6pt}
        \subcaption{SDARTS-RS~\cite{sdarts}}
\resizebox{\textwidth}{18mm}{
\begin{tabular}{ccccccc}
\hline
Partial data & C10 val     & C10 test    & C100 val    & C100 test   & Img val    & Img test   \\ \hline
100\%        & 72.90±4.00  & 77.35±5.57  & 44.53±5.16  & 44.94±5.17  & 24.82±5.15 & 23.66±4.55 \\
75\%         & 72.90±4.00  & 77.35±5.57  & 44.53±5.16  & 44.94±5.17  & 24.82±5.15 & 23.66±4.55 \\
50\%         & 72.90±4.00  & 77.35±5.57  & 44.53±5.16  & 44.94±5.17  & 24.82±5.15 & 23.66±4.55 \\
25\%         & 75.21±0.00  & 80.57±0.00  & 47.51±0.00  & 47.93±0.00  & 27.79±0.00 & 26.29±0.00 \\
20\%         & 63.40±20.46 & 71.81±15.17 & 36.68±18.75 & 37.16±18.66 & 24.00±6.56 & 22.97±5.76 \\
10\%         & 39.77±0.00  & 54.30±0.00  & 15.03±0.00  & 15.61±0.00  & 16.43±0.00 & 16.32±0.00 \\
5\%          & 39.77±0.00  & 54.30±0.00  & 15.03±0.00  & 15.61±0.00  & 16.43±0.00 & 16.32±0.00 \\
2\%          & 39.77±0.00  & 54.30±0.00  & 15.03±0.00  & 15.61±0.00  & 16.43±0.00 & 16.32±0.00 \\
1\%          & 39.77±0.00  & 54.30±0.00  & 15.03±0.00  & 15.61±0.00  & 16.43±0.00 & 16.32±0.00 \\
optimal      & 91.61       & 94.37       & 73.49       & 73.51       & 46.77      & 47.31      \\ \hline
\end{tabular}}
	\end{minipage}
	\hfill
	\begin{minipage}{0.48\linewidth}
	\centering
        \vspace{6pt}
        \subcaption{DARTS-flooding}
\resizebox{\textwidth}{18mm}{
\begin{tabular}{ccccccc}
\hline
Partial data & C10 val     & C10 test    & C100 val    & C100 test   & Img val     & Img test    \\ \hline
100\%        & 83.19±6.99  & 86.94±6.18  & 59.97±10.65 & 60.03±10.57 & 31.56±11.23 & 30.81±12.21 \\
75\%         & 71.77±5.01  & 75.64±5.58  & 45.25±4.01  & 45.43±4.36  & 17.15±3.36  & 16.34±2.68  \\
50\%         & 79.14±11.59 & 83.06±12.09 & 56.19±13.71 & 56.27±13.90 & 25.11±9.69  & 25.09±10.07 \\
25\%         & 82.04±8.83  & 85.88±8.78  & 59.77±11.06 & 59.57±11.31 & 30.50±11.91 & 30.19±12.77 \\
20\%         & 82.20±8.85  & 85.92±8.86  & 59.44±10.82 & 59.24±10.77 & 29.97±11.63 & 29.39±11.86 \\
10\%         & 83.23±7.25  & 87.00±5.89  & 59.03±10.43 & 58.90±10.07 & 33.80±6.15  & 33.15±6.95  \\
5\%          & 63.40±20.46 & 71.81±15.17 & 36.68±18.75 & 37.16±18.66 & 24.00±6.56  & 22.97±5.76  \\
2\%          & 51.58±20.46 & 63.06±15.17 & 25.86±18.75 & 26.38±18.66 & 20.22±6.56  & 19.64±5.76  \\
1\%          & 39.77±0.00  & 54.30±0.00  & 15.03±0.00  & 15.61±0.00  & 16.43±0.00  & 16.32±0.00  \\
optimal      & 91.61       & 94.37       & 73.49       & 73.51       & 46.77       & 47.31       \\ \hline
\end{tabular}}
	\end{minipage}
  	\qquad
  	\begin{minipage}{0.48\linewidth}
	\centering
        \vspace{6pt}
        \subcaption{$\beta$-DARTS~\cite{ye2022beta}}
\resizebox{\textwidth}{18mm}{
\begin{tabular}{ccccccc}
\hline
Partial data & C10 val    & C10 test   & C100 val   & C100 test  & Img val    & Img test   \\ \hline
100\%        & 91.55±0.00 & 94.36±0.00 & 73.49±0.00 & 73.51±0.00 & 46.37±0.00 & 46.34±0.00 \\
75\%         & 91.55±0.00 & 94.36±0.00 & 73.49±0.00 & 73.51±0.00 & 46.37±0.00 & 46.34±0.00 \\
50\%         & 91.55±0.00 & 94.36±0.00 & 73.49±0.00 & 73.51±0.00 & 46.37±0.00 & 46.34±0.00 \\
25\%         & 91.42±0.14 & 94.15±0.31 & 72.65±0.81 & 72.59±0.96 & 46.00±0.67 & 46.31±0.37 \\
20\%         & 90.76±0.29 & 93.44±0.23 & 70.32±0.47 & 70.25±0.73 & 43.89±0.18 & 44.16±0.56 \\
10\%         & 86.78±2.82 & 89.89±2.76 & 63.79±4.90 & 63.55±5.45 & 37.06±5.58 & 36.80±6.04 \\
5\%          & 88.36±3.09 & 91.24±2.92 & 66.03±5.70 & 66.12±6.22 & 39.85±5.49 & 39.37±6.26 \\
2\%          & 87.25±3.85 & 90.44±3.98 & 64.38±7.64 & 64.73±7.16 & 36.94±7.81 & 37.03±8.70 \\
1\%          & 88.21±1.10 & 91.11±1.26 & 67.97±1.58 & 67.84±1.58 & 40.90±2.44 & 40.80±2.23 \\
optimal      & 91.61      & 94.37      & 73.49      & 73.51      & 46.77      & 47.31      \\ \hline
\end{tabular}}
	\end{minipage}
	\hfill
	\begin{minipage}{0.48\linewidth}
	\centering
        \vspace{6pt}
        \subcaption{$\beta$-DARTS++}
\resizebox{\textwidth}{18mm}{
\begin{tabular}{ccccccc}
\hline
Partial data & C10 val    & C10 test   & C100 val   & C100 test  & Img val    & Img test   \\ \hline
100\%        & 91.55±0.00 & 94.36±0.00 & 73.49±0.00 & 73.51±0.00 & 46.37±0.00 & 46.34±0.00 \\
75\%         & 91.55±0.00 & 94.36±0.00 & 73.49±0.00 & 73.51±0.00 & 46.37±0.00 & 46.34±0.00 \\
50\%         & 91.55±0.00 & 94.36±0.00 & 73.49±0.00 & 73.51±0.00 & 46.37±0.00 & 46.34±0.00 \\
25\%         & 91.56±0.04 & 94.32±0.08 & 73.12±0.37 & 73.30±0.18 & 46.08±0.45 & 46.51±0.19 \\
20\%         & 91.55±0.07 & 94.28±0.15 & 72.88±0.56 & 72.98±0.69 & 46.14±0.51 & 46.14±0.69 \\
10\%         & 89.89±0.37 & 93.34±0.27 & 69.92±1.02 & 70.40±0.37 & 42.38±1.42 & 42.79±1.76 \\
5\%          & 89.49±0.28 & 93.29±0.02 & 70.85±0.13 & 71.09±0.12 & 41.47±0.23 & 42.48±0.68 \\
2\%          & 90.07±0.05 & 93.39±0.34 & 69.98±0.96 & 70.61±0.55 & 40.39±3.32 & 41.07±3.65 \\
1\%          & 90.34±0.19 & 93.20±0.28 & 69.81±0.34 & 70.14±0.15 & 45.02±0.14 & 45.47±0.17 \\
optimal      & 91.61      & 94.37      & 73.49      & 73.51      & 46.77      & 47.31      \\ \hline
\end{tabular}}
	\end{minipage}
\end{table*}

\begin{table*}
\caption{Performance comparison under different proxy channels (chan). We benchmark different methods including DARTS~\cite{darts}, RDARTS-L2~\cite{rdarts} and SDARTS-RS~\cite{sdarts}, and compare with our methods including DARTS-flooding, $\beta$-DARTS~\cite{ye2022beta} and $\beta$-DARTS++. All the results are averaged on 3 independent runs of searching.}
\label{table:proxy-chan}
    \centering
	\begin{minipage}{0.48\linewidth}
	\centering
        \subcaption{DARTS~\cite{darts}}
\resizebox{\textwidth}{12mm}{
\begin{tabular}{ccccccc}
\hline
Partial chan & C10 val    & C10 test   & C100 val    & C100 test   & Img val    & Img test   \\ \hline
16           & 39.77±0.00 & 54.30±0.00 & 15.03±0.00  & 15.61±0.00  & 16.43±0.00 & 16.32±0.00 \\
8            & 39.77±0.00 & 54.30±0.00 & 15.03±0.00  & 15.61±0.00  & 16.43±0.00 & 16.32±0.00 \\
4            & 70.05±3.05 & 72.91±3.44 & 40.18±2.78  & 40.70±3.00  & 20.00±1.96 & 19.27±1.48 \\
2            & 78.58±9.02 & 80.96±9.25 & 51.66±11.58 & 52.18±11.41 & 26.99±7.62 & 26.06±7.67 \\
1            & 88.53±1.35 & 91.78±1.39 & 66.73±3.09  & 66.93±2.64  & 38.19±3.42 & 38.12±3.54 \\
optimal      & 91.61      & 94.37      & 73.49       & 73.51       & 46.77      & 47.31      \\ \hline
\end{tabular}}
	\end{minipage}
	\hfill
	\begin{minipage}{0.48\linewidth}
	\centering
        \subcaption{RDARTS-L2~\cite{rdarts}.}
\resizebox{\textwidth}{12mm}{
\begin{tabular}{ccccccc}
\hline
Partial chan & C10 val     & C10 test   & C100 val    & C100 test   & Img val    & Img test   \\ \hline
16           & 58.78±16.47 & 65.38±9.60 & 30.72±13.59 & 31.18±13.49 & 18.06±1.41 & 17.71±1.21 \\
8            & 70.60±4.00  & 74.14±5.57 & 41.55±5.16  & 41.96±5.17  & 21.84±5.15 & 21.04±4.55 \\
4            & 78.53±5.74  & 83.37±4.84 & 52.34±8.36  & 52.48±7.88  & 29.80±3.48 & 28.67±4.12 \\
2            & 85.31±2.85  & 87.81±3.34 & 59.87±5.69  & 59.95±5.43  & 31.78±4.35 & 30.89±4.74 \\
1            & 89.01±1.52  & 92.05±1.27 & 68.10±1.41  & 68.43±1.16  & 40.84±1.67 & 41.62±2.41 \\
optimal      & 91.61       & 94.37      & 73.49       & 73.51       & 46.77      & 47.31      \\ \hline
\end{tabular}}
	\end{minipage}
 	\qquad
  	\begin{minipage}{0.48\linewidth}
	\centering
        \vspace{6pt}
        \subcaption{SDARTS-RS~\cite{sdarts}}
\resizebox{\textwidth}{12mm}{
\begin{tabular}{ccccccc}
\hline
Partial chan & C10 val    & C10 test   & C100 val   & C100 test  & Img val    & Img test   \\ \hline
16           & 72.90±4.00 & 77.35±5.57 & 44.53±5.16 & 44.94±5.17 & 24.82±5.15 & 23.66±4.55 \\
8            & 82.40±6.30 & 86.34±5.10 & 58.71±9.95 & 58.78±9.74 & 35.05±6.63 & 34.33±7.18 \\
4            & 90.32±0.98 & 93.22±1.04 & 70.02±2.28 & 70.27±2.25 & 42.86±2.33 & 43.23±3.71 \\
2            & 91.04±0.26 & 94.00±0.08 & 71.22±0.69 & 71.64±0.58 & 44.63±0.97 & 45.50±1.14 \\
1            & 86.76±2.27 & 89.77±2.89 & 64.65±6.02 & 64.91±5.72 & 36.92±6.78 & 36.32±7.17 \\
optimal      & 91.61      & 94.37      & 73.49      & 73.51      & 46.77      & 47.31      \\ \hline
\end{tabular}}
	\end{minipage}
	\hfill
	\begin{minipage}{0.48\linewidth}
	\centering
        \vspace{6pt}
        \subcaption{DARTS-flooding}
\resizebox{\textwidth}{12mm}{
\begin{tabular}{ccccccc}
\hline
Partial chan & C10 val     & C10 test    & C100 val    & C100 test   & Img val     & Img test    \\ \hline
16           & 75.83±8.99  & 79.74±9.14  & 49.77±10.19 & 50.34±10.39 & 22.54±9.06  & 22.09±9.72  \\
8            & 84.38±0.59  & 88.28±0.22  & 61.25±0.58  & 62.14±0.40  & 29.85±0.44  & 29.71±0.92  \\
4            & 79.32±10.58 & 82.99±11.37 & 55.89±12.94 & 55.67±12.56 & 28.79±12.15 & 28.41±13.05 \\
2            & 87.58±2.10  & 91.00±1.77  & 65.57±3.44  & 65.70±4.11  & 36.02±5.01  & 36.30±5.06  \\
1            & 89.52±0.90  & 92.86±0.86  & 68.44±1.49  & 69.07±1.54  & 41.30±2.21  & 41.30±2.45  \\
optimal      & 91.61       & 94.37       & 73.49       & 73.51       & 46.77       & 47.31       \\ \hline
\end{tabular}}
	\end{minipage}
  	\qquad
  	\begin{minipage}{0.48\linewidth}
	\centering
        \vspace{6pt}
        \subcaption{$\beta$-DARTS~\cite{ye2022beta}}
\resizebox{\textwidth}{12mm}{
\begin{tabular}{ccccccc}
\hline
Partia chan & C10 val    & C10 test   & C100 val   & C100 test  & Img val    & Img test   \\ \hline
16          & 91.55±0.00 & 94.36±0.00 & 73.49±0.00 & 73.51±0.00 & 46.37±0.00 & 46.34±0.00 \\
8           & 91.55±0.00 & 94.36±0.00 & 73.49±0.00 & 73.51±0.00 & 46.37±0.00 & 46.34±0.00 \\
4           & 90.65±0.78 & 93.96±0.35 & 71.64±1.61 & 71.91±1.39 & 42.64±3.23 & 43.07±2.83 \\
2           & 90.29±0.16 & 93.56±0.19 & 71.01±0.80 & 71.22±0.71 & 41.98±2.30 & 42.89±3.15 \\
1           & 90.34±0.90 & 93.76±0.50 & 70.51±2.16 & 70.59±1.82 & 44.01±2.27 & 44.30±2.15 \\
optimal     & 91.61      & 94.37      & 73.49      & 73.51      & 46.77      & 47.31      \\ \hline
\end{tabular}}
	\end{minipage}
	\hfill
	\begin{minipage}{0.48\linewidth}
	\centering
        \vspace{6pt}
        \subcaption{$\beta$-DARTS++}
\resizebox{\textwidth}{12mm}{
\begin{tabular}{ccccccc}
\hline
Partia chan & C10 val    & C10 test   & C100 val   & C100 test  & Img val    & Img test   \\ \hline
16          & 91.55±0.00 & 94.36±0.00 & 73.49±0.00 & 73.51±0.00 & 46.37±0.00 & 46.34±0.00 \\
8           & 91.55±0.00 & 94.36±0.00 & 73.49±0.00 & 73.51±0.00 & 46.37±0.00 & 46.34±0.00 \\
4           & 91.08±0.77 & 94.16±0.35 & 72.50±1.56 & 72.57±1.28 & 44.25±3.03 & 44.70±2.83 \\
2           & 90.65±0.78 & 93.96±0.35 & 71.64±1.61 & 71.91±1.39 & 42.64±3.23 & 43.07±2.83 \\
1           & 90.60±0.76 & 93.68±0.59 & 71.39±1.25 & 71.42±1.17 & 44.11±2.30 & 44.08±2.48 \\
optimal     & 91.61      & 94.37      & 73.49      & 73.51      & 46.77      & 47.31      \\ \hline
\end{tabular}}
	\end{minipage}
\end{table*}

\begin{table*}
\centering
\caption{Performance comparison under different proxy layers. We benchmark different methods including DARTS~\cite{darts}, RDARTS-L2~\cite{rdarts} and SDARTS-RS~\cite{sdarts}, and compare with our methods including DARTS-flooding, $\beta$-DARTS~\cite{ye2022beta} and $\beta$-DARTS++. All the results are averaged on 3 independent runs of searching.}
\label{table:proxy-layers}
	\begin{minipage}{0.48\linewidth}
	\centering
        \subcaption{DARTS~\cite{darts}}
\resizebox{\textwidth}{12mm}{
\begin{tabular}{ccccccc}
\hline
Partial layer & C10 val     & C10 test    & C100 val    & C100 test   & Img val    & Img test   \\ \hline
5             & 39.77±0.00  & 54.30±0.00  & 15.03±0.00  & 15.61±0.00  & 16.43±0.00 & 16.32±0.00 \\
4             & 39.77±0.00  & 54.30±0.00  & 15.03±0.00  & 15.61±0.00  & 16.43±0.00 & 16.32±0.00 \\
3             & 39.77±0.00  & 54.30±0.00  & 15.03±0.00  & 15.61±0.00  & 16.43±0.00 & 16.32±0.00 \\
2             & 39.77±0.00  & 54.30±0.00  & 15.03±0.00  & 15.61±0.00  & 16.43±0.00 & 16.32±0.00 \\
1             & 62.27±19.48 & 69.33±13.02 & 34.05±16.47 & 34.77±16.59 & 20.24±3.30 & 19.42±2.68 \\
optimal       & 91.61       & 94.37       & 73.49       & 73.51       & 46.77      & 47.31      \\ \hline
\end{tabular}}
	\end{minipage}
	\hfill
	\begin{minipage}{0.48\linewidth}
	\centering
        \subcaption{RDARTS-L2~\cite{rdarts}}
\resizebox{\textwidth}{12mm}{
\begin{tabular}{ccccccc}
\hline
Partial layer & C10 val     & C10 test   & C100 val    & C100 test   & Img val    & Img test   \\ \hline
5             & 58.78±16.47 & 65.38±9.60 & 30.72±13.59 & 31.18±13.49 & 18.06±1.41 & 17.71±1.21 \\
4             & 83.43±1.41  & 85.51±2.31 & 56.58±3.02  & 56.50±2.92  & 29.34±4.02 & 28.59±4.17 \\
3             & 89.17±0.18  & 92.00±0.09 & 66.64±0.36  & 66.66±0.33  & 39.95±0.31 & 39.33±0.44 \\
2             & 89.49±0.86  & 92.47±0.56 & 67.35±2.31  & 67.46±1.79  & 39.71±3.68 & 40.07±2.99 \\
1             & 87.58±0.00  & 91.06±0.00 & 66.05±0.00  & 65.61±0.00  & 27.83±0.00 & 27.78±0.00 \\
optimal       & 91.61       & 94.37      & 73.49       & 73.51       & 46.77      & 47.31      \\ \hline
\end{tabular}}
	\end{minipage}
 	\qquad
  	\begin{minipage}{0.48\linewidth}
	\centering
        \vspace{6pt}
        \subcaption{SDARTS-RS~\cite{sdarts}}
\resizebox{\textwidth}{12mm}{
\begin{tabular}{ccccccc}
\hline
Partial layer & C10 val     & C10 test   & C100 val    & C100 test   & Img val    & Img test   \\ \hline
5             & 72.90±4.00  & 77.35±5.57 & 44.53±5.16  & 44.94±5.17  & 24.82±5.15 & 23.66±4.55 \\
4             & 72.90±4.00  & 77.35±5.57 & 44.53±5.16  & 44.94±5.17  & 24.82±5.15 & 23.66±4.55 \\
3             & 58.78±16.47 & 65.38±9.60 & 30.72±13.59 & 31.18±13.49 & 18.06±1.41 & 17.71±1.21 \\
2             & 39.77±0.00  & 54.30±0.00 & 15.03±0.00  & 15.61±0.00  & 16.43±0.00 & 16.32±0.00 \\
1             & 39.77±0.00  & 54.30±0.00 & 15.03±0.00  & 15.61±0.00  & 16.43±0.00 & 16.32±0.00 \\
optimal       & 91.61       & 94.37      & 73.49       & 73.51       & 46.77      & 47.31      \\ \hline
\end{tabular}}
	\end{minipage}
	\hfill
	\begin{minipage}{0.48\linewidth}
	\centering
        \vspace{6pt}
        \subcaption{DARTS-flooding}
\resizebox{\textwidth}{12mm}{
\begin{tabular}{ccccccc}
\hline
Partial layer & C10 val     & C10 test    & C100 val    & C100 test   & Img val     & Img test    \\ \hline
5             & 75.83±8.99  & 79.74±9.14  & 49.77±10.19 & 50.34±10.39 & 22.54±9.06  & 22.09±9.72  \\
4             & 77.23±12.47 & 81.78±11.26 & 54.13±12.49 & 54.20±13.02 & 25.46±10.12 & 25.58±10.65 \\
3             & 69.88±9.31  & 74.20±9.01  & 46.05±9.73  & 46.09±9.59  & 15.04±2.18  & 14.63±1.75  \\
2             & 86.04±0.96  & 89.87±0.76  & 62.88±1.30  & 63.05±1.67  & 34.75±2.61  & 34.70±3.12  \\
1             & 79.63±7.64  & 83.78±7.87  & 56.11±9.30  & 56.46±9.42  & 25.16±9.81  & 24.97±9.48  \\
optimal       & 91.61       & 94.37       & 73.49       & 73.51       & 46.77       & 47.31       \\ \hline
\end{tabular}}
	\end{minipage}
  	\qquad
  	\begin{minipage}{0.48\linewidth}
	\centering
        \vspace{6pt}
        \subcaption{$\beta$-DARTS~\cite{ye2022beta}}
\resizebox{\textwidth}{12mm}{
\begin{tabular}{ccccccc}
\hline
Partial layer & C10 val    & C10 test   & C100 val   & C100 test  & Img val    & Img test   \\ \hline
5             & 91.55±0.00 & 94.36±0.00 & 73.49±0.00 & 73.51±0.00 & 46.37±0.00 & 46.34±0.00 \\
4             & 91.55±0.00 & 94.36±0.00 & 73.49±0.00 & 73.51±0.00 & 46.37±0.00 & 46.34±0.00 \\
3             & 91.55±0.00 & 94.36±0.00 & 73.49±0.00 & 73.51±0.00 & 46.37±0.00 & 46.34±0.00 \\
2             & 90.20±0.00 & 93.76±0.00 & 70.71±0.00 & 71.11±0.00 & 40.78±0.00 & 41.44±0.00 \\
1             & 90.20±0.00 & 93.76±0.00 & 70.71±0.00 & 71.11±0.00 & 40.78±0.00 & 41.44±0.00 \\
optimal       & 91.61      & 94.37      & 73.49      & 73.51      & 46.77      & 47.31      \\ \hline
\end{tabular}}
	\end{minipage}
	\hfill
	\begin{minipage}{0.48\linewidth}
	\centering
        \vspace{6pt}
        \subcaption{$\beta$-DARTS++}
\resizebox{\textwidth}{12mm}{
\begin{tabular}{ccccccc}
\hline
Partial layer & C10 val    & C10 test   & C100 val   & C100 test  & Img val    & Img test   \\ \hline
5             & 91.55±0.00 & 94.36±0.00 & 73.49±0.00 & 73.51±0.00 & 46.37±0.00 & 46.34±0.00 \\
4             & 91.55±0.00 & 94.36±0.00 & 73.49±0.00 & 73.51±0.00 & 46.37±0.00 & 46.34±0.00 \\
3             & 91.55±0.00 & 94.36±0.00 & 73.49±0.00 & 73.51±0.00 & 46.37±0.00 & 46.34±0.00 \\
2             & 90.65±0.78 & 93.96±0.35 & 71.64±1.61 & 71.91±1.39 & 42.64±3.23 & 43.07±2.83 \\
1             & 90.20±0.00 & 93.76±0.00 & 70.71±0.00 & 71.11±0.00 & 40.78±0.00 & 41.44±0.00 \\
optimal       & 91.61      & 94.37      & 73.49      & 73.51      & 46.77      & 47.31      \\ \hline
\end{tabular}}
	\end{minipage}
\end{table*}

\subsection{The Robustness of Different NAS Methods under Various Proxies} \label{sec: proxy robustness}
In this section, we fully evaluate the search robustness of both commonly-used differentiable NAS methods (e.g., DARTS~\cite{darts}, SDARTS-RS~\cite{sdarts}, and RDARTS-L2~\cite{rdarts}) and our proposed methods (i.e., DARTS with Beta-Decay, weight flooding, and bi-level regularization) under a wide range of proxy data, proxy channels, proxy layers, and proxy epochs. Note that, all experiments are conducted on the most widely used NAS benchmark, namely NAS-Bench-201, and the search settings are kept the same as DARTS on~\cite{dong2020bench} except for changing various proxies.

\subsubsection{Robustness under proxy data} The search robustness of different methods under a wide range of proxy data is shown in Table.~\ref{table:proxy-data}. As we can see, reducing the proxy data degrades the performance of different NAS consistently. Especially, the performance of RDARTS-L2, SDARTS-RS and DARTS-flooding will degrade to the performance of DARTS when randomly sampling 25\%, 10\% and 1\% proxy data, respectively. As a comparison, our $\beta$-DARTS always obtains a SOTA performance when randomly sampling 100\% to 25\% proxy data. Although the performance suffers from degradation when randomly sampling 25\% to 1\% proxy data, $\beta$-DARTS still remains a comparable performance and performs much better than all other methods. In addition, by applying the proposed weight flooding regularization to $\beta$-DARTS (termed as $\beta$-DARTS++), the search performance under 25\% to 1\% proxy data can be improved significantly. Besides, DARTS-flooding shows better results than DARTS under kinds of proxy data, which further demonstrates the effectiveness of weight flooding regularization.

\subsubsection{Robustness under proxy channels} The search robustness of different methods under a wide range of proxy channels is shown in Table.~\ref{table:proxy-chan}. As we can see, reducing the proxy channels improves the performance of DARTS, RDARTS-L2, SDARTS-RS, and DARTS-flooding to some extent. On the one hand, such results verify that commonly-used proxy settings may not be optimal. On the other hand, such results also give some experimental evidences for the effectiveness of partial connected channel proposed in~\cite{pc-darts, xu2021partially}. As a comparison, our $\beta$-DARTS always reaches a SOTA performance when the proxy channels are set from 16 to 8. Although the performance is somewhat degraded when the proxy channels are set from 4 to 1, $\beta$-DARTS still maintains a comparable performance and performs much better than other methods under almost all settings. Further, by applying weight flooding regularization to $\beta$-DARTS (termed as $\beta$-DARTS++), the results under proxy channel 4 to 1 can be boosted consistently. Besides, DARTS-flooding always performs much better than original DARTS under kinds of proxy channels, which also verifies the effectiveness of weight flooding regularization.

\subsubsection{Robustness under proxy layers} The search robustness of different methods under a wide range of proxy layers is shown in Table.~\ref{table:proxy-layers}. As we can see, reducing the proxy layers improves the performance of RDARTS-L2, degrades the performance of SDARTS-RS, and may either improve or degrade the performance of DARTS-flooding. Such results verify that different proxies have different impact on different NAS methods. As a comparison, our $\beta$-DARTS always achieves a SOTA result when setting the proxy layers from 5 to 3. Although the performance is degraded  when setting the proxy layers to 2 and 1, $\beta$-DARTS still has a better result than all other methods. Further, by applying weight flooding regularization to DARTS and $\beta$-DARTS (termed as DARTS-flooding and $\beta$-DARTS++), both search performance can be improved, verifying the effectiveness of weight flooding regularization. 

\subsubsection{Robustness under proxy epochs} The search trajectories of different NAS methods are shown in Fig.~\ref{fig:3} and Fig.~\ref{fig:beta_variants}. As we can see, for all other methods including DARTS, RDARTS-L2, SDARTS-RS and SNAS, the last epoch is not necessarily the optimal epoch and the performance obtained by different epochs may vary greatly. As a comparison, for all the $\beta$-variants with the proposed Beta-Decay regularization, the last epoch is generally the optimal epoch and the performance is always optimal from about 20 epoch to 100 epoch.

\begin{table}[t]
\begin{center}
 \caption{The optimal epoch and corresponding search performance indicated by different criteria.} 
\label{tab:adaptive-epoch}
\begin{tabular}{ccc}
\hline
\multirow{2}{*}{\begin{tabular}[c]{@{}c@{}}Partial \\ data\end{tabular}} & \multicolumn{2}{c}{Criteria 1/2/3/4}   \\ \cline{2-3} 
& End epoch    & C10 val                 \\ \hline
100\%                                                                    & 8/13/50/100  & 91.55/91.55/91.55/91.55 \\
75\%                                                                     & 10/27/31/100 & 91.55/91.55/91.55/91.55 \\
50\%                                                                     & 8/9/34/100   & 91.55/91.55/91.55/91.55 \\
25\%                                                                     & 5/12/25/100  & 90.98/91.55/91.55/91.55 \\
20\%                                                                     & 7/9/20/100   & 91.48/91.48/91.48/91.48 \\
10\%                                                                     & 5/9/31/100   & 89.76/90.30/90.30/90.02 \\
5\%                                                                      & 2/6/11/100   & 89.55/91.07/91.35/89.65 \\
2\%                                                                      & 1/3/6/100    & 88.79/90.11/90.11/90.11 \\
1\%                                                                      & 1/2/7/100    & 88.10/88.65/90.45/90.45 \\ \hline
\end{tabular}
\end{center}
\vspace{-12pt}
\end{table}

On the other hand, since all the $\beta$-variants reach the optimal point at an early epoch, searching until the last epoch consumes both time and computation. Thus, a criteria that can automatically indicate the optimal epoch is desired. Further, observing Fig.~\ref{fig:2}(c), we can find that, when the alpha standard deviation of different edges of $\beta$-DARTS increases to a certain extent, it will remain unchanged until the last epoch, which is highly correlated to the search trajectory in Fig.~\ref{fig:3}. Inspired by this, we attempt to use the alpha standard deviation to design the desired criteria in our $\beta$-DARTS++. Naturally, when the alpha standard deviation of a certain edge $(i, j)$ no longer increases, we can consider that the operator rank of this edge has been determined. Suppose there are total $M$ edges, and $m$ edges have been determined, we define three optional criteria as follows
\begin{equation} \label{eqn17}
\small
\text { End the search, if }\left\{\begin{array}{l}
m=1, ~~\text { Criteria 1} \\
m \geq \frac{M}{2}, \text { Criteria 2} \\
m=M, \,\text { Criteria 3}
\end{array}\right.
  \vspace{-6pt}
\end{equation}

For \bp{an} intuitive comparison, we also include searching until the last epoch as criteria 4. The comparison results are shown in Table~\ref{tab:adaptive-epoch}. As we can see, all the proposed criteria 1, 2 and 3 can automatically decide the optimal epoch of searching under different proxy data. And from criteria 1 to criteria 2 to criteria 3, the end epoch becomes larger, while the search performance usually becomes better. In particular, criteria 2 and criteria 3 can achieve almost the same performance with that of searching until the last epoch, while using much less epochs, verifying the effectiveness and efficiency of the alpha standard deviation based criteria. More similar results can be found in \bp{Appendix B.2}.


\subsection{Ablation study}
\noindent\textbf{The Importance of Increased Weighting Scheme for $\lambda$.} We first explore the influence of different weighting schemes on DARTS with Beta-Decay regularization, including linear increased weighting scheme, constant weighting scheme and linear decay weighting scheme. The results are shown in Table~\ref{tab:3}. As we can see, the linear decay weighting scheme impedes the effect of regularization, the constant weighting scheme is sensitive to the hyperparameter, while the linear increased weighting scheme is not only effective but also robust to the hyperparameter. Besides, combining with the results of Fig.~\ref{fig:3} that the performance on CIFAR-10, CIFAR-100, and Imagenet in a single run of searching reach the optimal point sequentially, we could conclude that the linear increased regularization coefficient can further improve the generalization ability of inferred model after the searching performance on current data is maximized, as evidenced by Eq.~(\ref{eqn12}) and Eq.~(\ref{eqn16}).

\begin{figure}[t] 
    \vspace{-4pt}	
	\centering
	\includegraphics[width=3.3in]{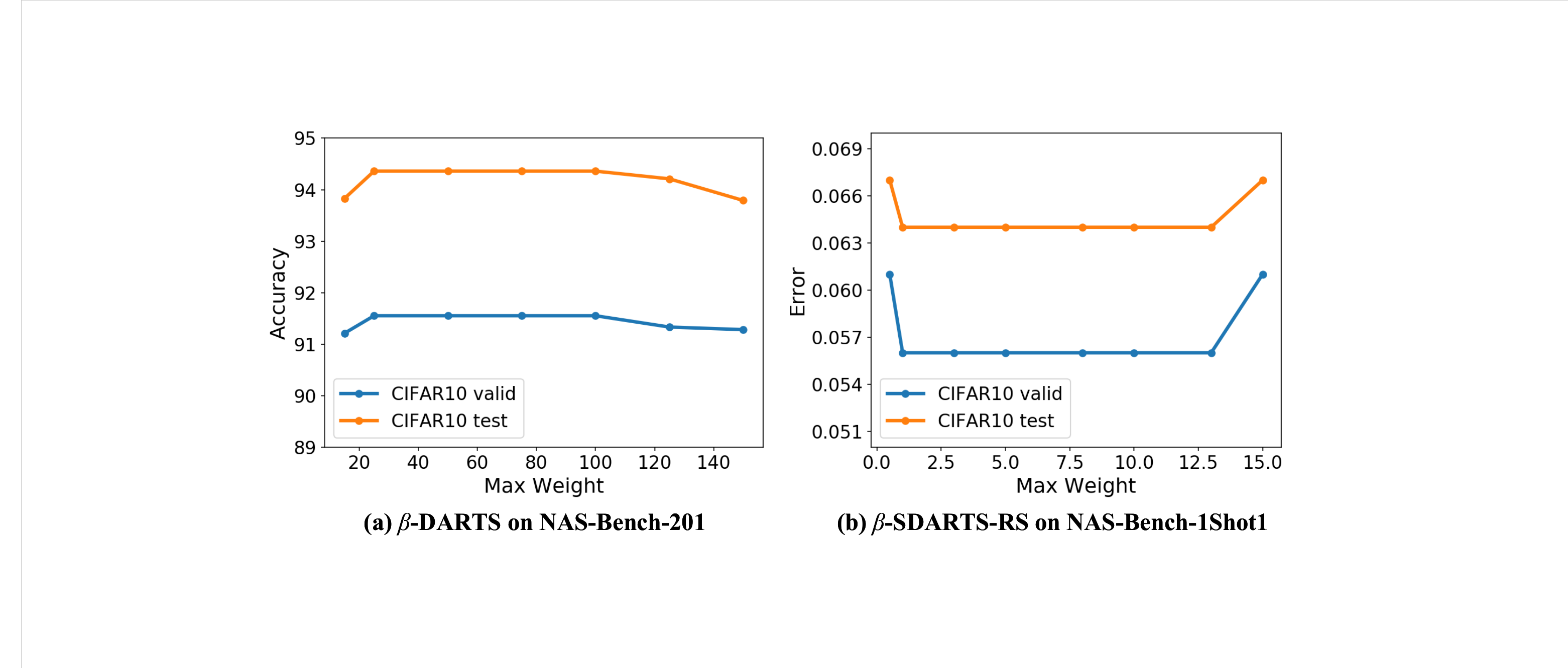}
	\vspace{-6pt}	
	\caption{The effect of different max weight of linear increased weighting schemes on the searching results of the proposed Beta-Decay regularization.}
	\label{fig:5}
	\vspace{-12pt}	
\end{figure}

\noindent\textbf{Wide Range of The Optimal Weight for $\lambda$.} We further investigate the optimal \bp{maximal} weight of the linear increased weighting scheme of Beta-Decay regularization. The results on CIFAR-10 of NAS-Bench-201 and search space 1 of NAS-Bench-1Shot1 are provided in Fig.~\ref{fig:5}. We can see that the best performance is achieved in a wide range of \bp{maximal} weights, namely about 25-100 and 1-13 for $\beta$-DARTS on NAS-Bench-201 and $\beta$-SDARTS-RS on NAS-Bench-1Shot1 respectively. There are similar results on CIFAR-10 and CIFAR-100 in common DARTS search space, as shown in \bp{Appendix B.3}. If not mentioned specially, the default values of \bp{maximal} weight for NAS-Bench-201, NAS-Bench-1Shot1, DARTS-CIFAR-10,DARTS-CIFAR-100, and MobileNet search space are set to 50, 7, 0.5, 5, and 20 respectively in all our experiments. Furthermore, comparing Eq.~(\ref{eqn12}) with Eq.~(\ref{eqn5}) and Eq.~(\ref{eqn7}), we find that the normalized values of $\alpha$ in  Eq.~(\ref{eqn12}) has the ability to make sure that the optimization process is not sensitive to the hyperparameter of $\lambda$.

\begin{table}[t]
\begin{center}
 \caption{Influence of different weighting schemes on the proposed Beta-Decay regularization.} 
\label{tab:3}
\resizebox{\linewidth}{!}{
\begin{tabular}{lcc}
\hline
Weighting Scheme & CIFAR-10 valid          & CIFAR-10 test           \\ \hline
0-15/25/50/100   & 91.21/91.55/91.55/91.55 & 93.83/94.36/94.36/94.36 \\
5/10/15/25     & 84.96/90.59/91.55/90.59        & 88.02/93.31/94.36/93.31      \\
25-15/10/5/0   & 90.59/87.30/73.58/39.77        & 93.31/90.65/76.88/54.30      \\ \hline
\end{tabular}
}
\end{center}
\vspace{-6pt}
\end{table}
\begin{table*}[t]
\begin{center}
 \caption{Influence of different coefficient $b$ of Flooding regularization. All the results are averaged on 3 independent runs of searching with 20\% proxy data on the NAS-Bench-201 benchmark.}
\label{tab:flooding-coef}
\begin{tabular}{ccccccccc}
\hline
\multirow{2}{*}{\begin{tabular}[c]{@{}c@{}}Partial \\ data\end{tabular}} & \multirow{2}{*}{\begin{tabular}[c]{@{}c@{}}Flooding \\ Coefficient\end{tabular}} & \multirow{2}{*}{\begin{tabular}[c]{@{}c@{}}Weighting\\ Scheme\end{tabular}} & \multicolumn{2}{c}{CIFAR-10} & \multicolumn{2}{c}{CIFAR-100} & \multicolumn{2}{c}{ImageNet16-120} \\ \cline{4-9} 
&                                                                                  &                                                                             & valid         & test         & valid         & test          & valid            & test            \\ \hline
20\%                                                                     & 0                                                                                & 0-50                                                                        & 90.76±0.29    & 93.44±0.23   & 70.32±0.47    & 70.25±0.73    & 43.89±0.18       & 44.16±0.56      \\
20\%                                                                     & 0.5                                                                              & 0-50                                                                        & 91.44±0.17    & 94.15±0.31   & 72.41±0.46    & 72.50±0.82    & 45.73±0.61       & 46.44±0.44      \\
20\%                                                                     & 0.8                                                                              & 0-50                                                                        & 91.47±0.07    & 94.33±0.04   & 72.94±0.48    & 72.98±0.46    & 46.05±0.73       & 46.29±0.33      \\
20\%                                                                     & 1.0                                                                              & 0-50                                                                        & 91.55±0.07    & 94.28±0.15   & 72.88±0.56    & 72.98±0.69    & 46.14±0.51       & 46.14±0.69      \\
20\%                                                                     & 1.2                                                                              & 0-50                                                                        & 91.53±0.10    & 94.34±0.04   & 72.94±0.48    & 73.13±0.43    & 45.72±0.59       & 46.33±0.39      \\
20\%                                                                     & 1.5                                                                              & 0-50                                                                        & 91.48±0.06    & 94.29±0.08   & 73.01±0.37    & 72.98±0.27    & 45.71±0.56       & 46.25±0.28      \\ \hline
\end{tabular}
\end{center}
\vspace{-5pt}
\end{table*}

\noindent\textbf{Wide Range of The Optimal Flooding Coefficient $b$.} We also study the influence of different flooding regularization coefficients. The results of searching with 20\% proxy data on NAS-Bench-201 benchmark are shown in Table~\ref{tab:flooding-coef}. We can see that, the performance of different datasets of $\beta$-DARTS with reduced proxy data can be greatly improved by flooding regularization, and the improvement is achieved over a wide range of regularization coefficients. The reason may be that different regularization coefficients always hit the target, pushing the network weights to a more robust area with a smoother parameter landscape, as demonstrated in Eq.~\ref{descent_ascent_1}. For simplicity, we uniformly set the flooding regularization coefficients to 1.0, 1.0, and 0.5 for various proxy data, various proxy channels, and various proxy layers. 

\begin{table*}[t]
\begin{center}
\caption{The results of different Beta regularization loss with different weighting schemes on NAS-Bench-201 benchmark. Note that we only search on CIFAR-10 dataset, and perform 2 runs of searching under different random seeds.}
\label{tab:4}
\small
\begin{tabular}{lccccccc}
\hline
\multirow{2}{*}{Methods} & \multirow{2}{*}{\begin{tabular}[c]{@{}c@{}}Weighting\\ Scheme\end{tabular}} & \multicolumn{2}{c}{CIFAR-10} & \multicolumn{2}{c}{CIFAR-100} & \multicolumn{2}{c}{ImageNet16-120} \\ \cline{3-8} 
                         &                                                                             & valid         & test         & valid         & test          & valid            & test            \\ \hline
DARTS(1st)~\cite{darts}               & 3.2                                                                     & 39.77±0.00    & 54.30±0.00   & 15.03±0.00    & 15.61±0.00    & 16.43±0.00       & 16.32±0.00     \\
Beta-Global                & 0-25                                                                        & 91.55/91.55   & 94.36/94.36  & 73.49/73.49   & 73.51/73.51   & 46.37/46.37      & 46.34/46.34     \\
Beta-Global                & 0-50                                                                        & 91.55/91.55   & 94.36/94.36  & 73.49/73.49   & 73.51/73.51   & 46.37/46.37      & 46.34/46.34     \\
Beta-Global                & 0-75                                                                        & 91.55/91.55   & 94.36/94.36  & 73.49/73.49   & 73.51/73.51   & 46.37/46.37      & 46.34/46.34     \\
Beta-Global                & 0-100                                                                       & 91.21/91.55   & 93.83/94.36  & 71.60/73.49   & 71.88/73.51   & 45.75/46.37      & 44.65/46.34     \\
Beta-Zero                  & 0-25                                                                        & 91.21/90.97   & 93.83/93.91  & 71.60/70.41   & 71.88/70.78   & 45.75/43.77      & 44.65/44.78                 \\
Beta-Zero                  & 0-50                                                                        & 91.55/91.21   & 94.36/93.83  & 73.49/71.60   & 73.51/71.88   & 46.37/45.74      & 46.34/44.65     \\
Beta-Zero                  & 0-75                                                                        & 91.61/91.05   & 94.37/93.66  & 72.75/71.02   & 73.22/71.38   & 45.56/45.23      & 46.71/44.70     \\
Beta-Zero                  & 0-100                                                                       & 91.21/91.21   & 93.83/93.83  & 71.60/71.60   & 71.88/71.88   & 45.75/45.75      & 44.65/44.65     \\ \hline
\end{tabular}
\end{center}
\vspace{-8pt}
\end{table*}

\begin{table}[t]
\begin{center}
 \caption{Effect of commonly-used weight regularization on $\beta$-DARTS under 20\% proxy data on NAS-Bench-201.} 
\label{tab:effect-weight-reg}
\begin{tabular}{lcc}
\hline
\multicolumn{1}{c}{Method} & C10 val           & C10 test          \\ \hline
BD                         & 91.10/90.59/90.59 & 93.71/93.31/93.31 \\
BD+RS-0.1                  & 91.07/89.65/89.97 & 93.99/93.30/93.07 \\
BD+RS-0.3                  & 90.31/91.07/89.48 & 93.33/93.99/92.81 \\
BD+RS-0.5                  & 89.65/90.49/90.31 & 93.30/93.65/93.33 \\
BD+L2-60e-4                & 90.66/90.68/90.66 & 93.28/93.45/93.28 \\
BD+L2-81e-4                & 89.96/90.31/91.34 & 92.57/93.33/94.00 \\
BD+L2-120e-4               & 90.66/91.14/90.66 & 93.28/93.74/93.28 \\
BD+FD-1.0                  & 91.55/91.61/91.48 & 94.36/94.37/94.11 \\
optimal                    & 91.61             & 94.37             \\ \hline
\end{tabular}
\end{center}
\vspace{-6pt}
\end{table}
\begin{table}[t]
\begin{center}
 \caption{Effect of asymmetric data partition on $\beta$-DARTS under 20\% proxy data on NAS-Bench-201.} 
\label{tab:asymmetric-data}
\begin{tabular}{ccc}
\hline
Partial data ($w/\alpha$) & C10 val           & C10 test          \\ \hline
15\%/5\%    & 91.33/91.21/90.86 & 94.21/93.83/93.72 \\
10\%/10\%      & 91.10/90.59/90.59 & 93.71/93.31/93.31 \\
5\%/15\%    & 85.13/87.61/89.76 & 88.28/90.22/92.54 \\ \hline
\end{tabular}
\end{center}
\vspace{-12pt}
\end{table}

\section{More Discussions and Findings } \label{sec:experiments-more}
\noindent\textbf{More Variations of Beta-Decay Regularization.} Actually, the idea of regularizing Beta is what really matters, and the way to realize it is non-unique. Here, we show two kinds of variants of Beta regularization loss. Recalling Eq.~(\ref{eqn10}), we can naturally figure out an alternative, using the $\operatorname{smoothmax}$ of all architecture parameters on the entire supernet, namely Beta-Global loss.
\begin{equation} \label{eqn17}
\small
  \begin{split}
    \mathcal{L}_{Beta-Global}&=\operatorname{smoothmax} \left(\begin{array}{r}\alpha_{1}^1,\cdots ,\alpha_{\left | \mathcal{O}\right |}^L\end{array}\right) \\
    &=\log \left(\sum_{l=1}^{L}\sum_{k=1}^{\left | \mathcal{O}\right |}  e^{\alpha_k^l}\right)
  \end{split}
  \vspace{-6pt}
\end{equation}

In addition, by introducing a threshold, we can get the $\operatorname{smoothmax}$ between the threshold and each architecture parameter. We simply set the threshold as 0 in this paper, namely Beta-Zero loss.
\begin{equation} \label{eqn18}
\small
  \begin{split}
    \mathcal{L}_{Beta-Zero}&=\operatorname{smoothmax} \left(\begin{array}{r}0, \alpha_{k}^l\end{array}\right) \\
    &= -\log \left(1+e^{-\alpha_k^l}\right)
  \end{split}
  \vspace{-6pt}
\end{equation}

The results of DARTS with Beta-Global and Beta-Zero regularization loss are shown in Table.~\ref{tab:4}. As we can see, both \bp{losses} can promote original DARTS by a large margin, while Beta-Global loss that takes the same effect with Beta-Decay loss, can more stably obtain better results than Beta-Zero loss under different weighting schemes. Such results validate that regularizing $\beta$ is important, while the way to achieve it has a lot of room for exploration.

\noindent\textbf{Commonly-used Weight Regularization May Not work Under Various Proxies.} We attempt to improve $\beta$-DARTS under reduced proxy using the weight regularization proposed in SDARTS~\cite{sdarts} and RDARTS~\cite{rdarts} as well. As shown in Table~\ref{tab:effect-weight-reg}, different random smoothing perturbations (RS) and different larger L2 regularization (L2) seem to have no impact on the search results, while flooding regularization (FD) can improve the performance significantly. The reason may be that constantly minimizing the loss value provides few guarantees on the weight robustness.

\noindent\textbf{Asymmetric Data Partition.} Since DARTS uses half of training data to optimize network weights and half of training data to optimize architecture parameters by default, we also explore the effect of asymmetric data partition. As shown in Table~\ref{tab:asymmetric-data}, we find that allocating network weights more \bp{training} data and architecture parameters less \bp{training} data seems to have a positive impact on the search performance.





\section{Conclusions} \label{sec:Con} 
In this paper, we investigate the explicit regularization on the optimization of DARTS in depth, which is \bp{generally} ignored by previous works. Firstly, we identify that L2 or weight decay regularization on architecture parameters alpha commonly used by DARTS and its variants may not be effective or even counterproductive. Then, we propose a novel and generic Beta-Decay regularization loss, for improving DARTS-based methods without extra changes or cost (termed as $\beta$-DARTS). We theoretically and experimentally show that Beta-Decay regularization can improve both the stability and the generalization of DARTS. In addition, we attempt to investigate the robustness of different NAS including $\beta$-DARTS under kinds of proxies for the first time. We conclude some interesting findings by benchmarking different NAS methods and further introduce a new weight flooding regularization into the NAS field. Theoretical analysis and extensive experiments show that weight flooding regularization can \bp{further} improve the robustness of $\beta$-DARTS under various proxies significantly.

\ifCLASSOPTIONcompsoc
  \section*{Acknowledgments}
\else
  \section*{Acknowledgment}
\fi
This work was supported by National Natural Science Foundation of China (No.62071127 and U1909207), and Zhejiang Lab Project (No. 2021KH0AB05). Wanli Ouyang was supported by the Australian Research Council Grant (DP200103223), Australian Medical Research Future Fund (MRFAI000085), and CRC-P Smart Material Recovery Facility SMRF-Curby Soft Plastics.

\ifCLASSOPTIONcaptionsoff
  \newpage
\fi



\bibliographystyle{IEEEtran}
\bibliography{main_paper}

\begin{thebibliography}{10}
\providecommand{\url}[1]{#1}
\csname url@samestyle\endcsname
\providecommand{\newblock}{\relax}
\providecommand{\bibinfo}[2]{#2}
\providecommand{\BIBentrySTDinterwordspacing}{\spaceskip=0pt\relax}
\providecommand{\BIBentryALTinterwordstretchfactor}{4}
\providecommand{\BIBentryALTinterwordspacing}{\spaceskip=\fontdimen2\font plus
\BIBentryALTinterwordstretchfactor\fontdimen3\font minus
  \fontdimen4\font\relax}
\providecommand{\BIBforeignlanguage}[2]{{%
\expandafter\ifx\csname l@#1\endcsname\relax
\typeout{** WARNING: IEEEtran.bst: No hyphenation pattern has been}%
\typeout{** loaded for the language `#1'. Using the pattern for}%
\typeout{** the default language instead.}%
\else
\language=\csname l@#1\endcsname
\fi
#2}}
\providecommand{\BIBdecl}{\relax}
\BIBdecl

\bibitem{nasnet}
B.~Zoph, V.~Vasudevan, J.~Shlens, and Q.~V. Le, ``Learning transferable
  architectures for scalable image recognition,'' in \emph{Proceedings of the
  IEEE conference on computer vision and pattern recognition}, 2018, pp.
  8697--8710.

\bibitem{mnasnet}
M.~Tan, B.~Chen, R.~Pang, V.~Vasudevan, M.~Sandler, A.~Howard, and Q.~V. Le,
  ``Mnasnet: Platform-aware neural architecture search for mobile,'' in
  \emph{Proceedings of the IEEE/CVF Conference on Computer Vision and Pattern
  Recognition}, 2019, pp. 2820--2828.

\bibitem{amoebanet}
E.~Real, A.~Aggarwal, Y.~Huang, and Q.~V. Le, ``Regularized evolution for image
  classifier architecture search,'' in \emph{Proceedings of the aaai conference
  on artificial intelligence}, vol.~33, no.~01, 2019, pp. 4780--4789.

\bibitem{klein2016learning}
A.~Klein, S.~Falkner, J.~T. Springenberg, and F.~Hutter, ``Learning curve
  prediction with bayesian neural networks,'' 2016.

\bibitem{cai2018path}
H.~Cai, J.~Yang, W.~Zhang, S.~Han, and Y.~Yu, ``Path-level network
  transformation for efficient architecture search,'' in \emph{International
  Conference on Machine Learning}.\hskip 1em plus 0.5em minus 0.4em\relax PMLR,
  2018, pp. 678--687.

\bibitem{spos}
Z.~Guo, X.~Zhang, H.~Mu, W.~Heng, Z.~Liu, Y.~Wei, and J.~Sun, ``Single path
  one-shot neural architecture search with uniform sampling,'' in
  \emph{European Conference on Computer Vision}.\hskip 1em plus 0.5em minus
  0.4em\relax Springer, 2020, pp. 544--560.

\bibitem{ye2022efficient}
P.~Ye, B.~Li, T.~Chen, J.~Fan, Z.~Mei, C.~Lin, C.~Zuo, Q.~Chi, and W.~Ouyang,
  ``Efficient joint-dimensional search with solution space regularization for
  real-time semantic segmentation,'' \emph{International Journal of Computer
  Vision}, vol. 130, no.~11, pp. 2674--2694, 2022.

\bibitem{ye2022stimulative}
P.~Ye, S.~Tang, B.~Li, T.~Chen, and W.~Ouyang, ``Stimulative training of
  residual networks: A social psychology perspective of loafing,'' \emph{arXiv
  preprint arXiv:2210.04153}, 2022.

\bibitem{darts}
H.~Liu, K.~Simonyan, and Y.~Yang, ``Darts: Differentiable architecture
  search,'' in \emph{International Conference on Learning Representations},
  2018.

\bibitem{darts-}
X.~Chu, X.~Wang, B.~Zhang, S.~Lu, X.~Wei, and J.~Yan, ``Darts-: Robustly
  stepping out of performance collapse without indicators,'' in
  \emph{International Conference on Learning Representations}, 2020.

\bibitem{snas}
S.~Xie, H.~Zheng, C.~Liu, and L.~Lin, ``Snas: stochastic neural architecture
  search,'' in \emph{International Conference on Learning Representations},
  2018.

\bibitem{rdarts}
T.~E. Arber~Zela, T.~Saikia, Y.~Marrakchi, T.~Brox, and F.~Hutter,
  ``Understanding and robustifying differentiable architecture search,'' in
  \emph{International Conference on Learning Representations}, vol.~3, 2020,
  p.~7.

\bibitem{sdarts}
X.~Chen and C.-J. Hsieh, ``Stabilizing differentiable architecture search via
  perturbation-based regularization,'' in \emph{International Conference on
  Machine Learning}.\hskip 1em plus 0.5em minus 0.4em\relax PMLR, 2020, pp.
  1554--1565.

\bibitem{zhou2020econas}
D.~Zhou, X.~Zhou, W.~Zhang, C.~C. Loy, S.~Yi, X.~Zhang, and W.~Ouyang,
  ``Econas: Finding proxies for economical neural architecture search,'' in
  \emph{Proceedings of the IEEE/CVF Conference on computer vision and pattern
  recognition}, 2020, pp. 11\,396--11\,404.

\bibitem{na2021accelerating}
B.~Na, J.~Mok, H.~Choe, and S.~Yoon, ``Accelerating neural architecture search
  via proxy data,'' \emph{arXiv preprint arXiv:2106.04784}, 2021.

\bibitem{pc-darts}
Y.~Xu, L.~Xie, X.~Zhang, X.~Chen, G.-J. Qi, Q.~Tian, and H.~Xiong, ``Pc-darts:
  Partial channel connections for memory-efficient architecture search,'' in
  \emph{International Conference on Learning Representations}, 2019.

\bibitem{xu2021partially}
Y.~Xu, L.~Xie, W.~Dai, X.~Zhang, X.~Chen, G.-J. Qi, H.~Xiong, and Q.~Tian,
  ``Partially-connected neural architecture search for reduced computational
  redundancy,'' \emph{IEEE Transactions on Pattern Analysis and Machine
  Intelligence}, vol.~43, no.~9, pp. 2953--2970, 2021.

\bibitem{ye2022beta}
P.~Ye, B.~Li, Y.~Li, T.~Chen, J.~Fan, and W.~Ouyang, ``$\beta $-darts:
  Beta-decay regularization for differentiable architecture search,''
  \emph{arXiv preprint arXiv:2203.01665}, 2022.

\bibitem{pdarts}
X.~Chen, L.~Xie, J.~Wu, and Q.~Tian, ``Progressive differentiable architecture
  search: Bridging the depth gap between search and evaluation,'' in
  \emph{Proceedings of the IEEE/CVF International Conference on Computer
  Vision}, 2019, pp. 1294--1303.

\bibitem{darts+}
H.~Liang, S.~Zhang, J.~Sun, X.~He, W.~Huang, K.~Zhuang, and Z.~Li, ``Darts+:
  Improved differentiable architecture search with early stopping,''
  \emph{arXiv preprint arXiv:1909.06035}, 2019.

\bibitem{fairdarts}
X.~Chu, T.~Zhou, B.~Zhang, and J.~Li, ``Fair darts: Eliminating unfair
  advantages in differentiable architecture search,'' in \emph{European
  conference on computer vision}.\hskip 1em plus 0.5em minus 0.4em\relax
  Springer, 2020, pp. 465--480.

\bibitem{dots}
Y.-C. Gu, L.-J. Wang, Y.~Liu, Y.~Yang, Y.-H. Wu, S.-P. Lu, and M.-M. Cheng,
  ``Dots: Decoupling operation and topology in differentiable architecture
  search,'' in \emph{Proceedings of the IEEE/CVF Conference on Computer Vision
  and Pattern Recognition}, 2021, pp. 12\,311--12\,320.

\bibitem{adaptNAS}
Y.~Li, Z.~Yang, Y.~Wang, and C.~Xu, ``Adapting neural architectures between
  domains,'' \emph{Advances in Neural Information Processing Systems}, vol.~33,
  2020.

\bibitem{mixsearch}
L.~Liu, Z.~Wen, S.~Liu, H.-Y. Zhou, H.~Zhu, W.~Xie, L.~Shen, K.~Ma, and
  Y.~Zheng, ``Mixsearch: Searching for domain generalized medical image
  segmentation architectures,'' \emph{arXiv preprint arXiv:2102.13280}, 2021.

\bibitem{cortes2012l2}
C.~Cortes, M.~Mohri, and A.~Rostamizadeh, ``L2 regularization for learning
  kernels,'' \emph{arXiv preprint arXiv:1205.2653}, 2012.

\bibitem{wd}
A.~Krogh and J.~A. Hertz, ``A simple weight decay can improve generalization,''
  in \emph{Advances in neural information processing systems}, 1992, pp.
  950--957.

\bibitem{ishida2020we}
T.~Ishida, I.~Yamane, T.~Sakai, G.~Niu, and M.~Sugiyama, ``Do we need zero
  training loss after achieving zero training error?'' in \emph{International
  Conference on Machine Learning}.\hskip 1em plus 0.5em minus 0.4em\relax PMLR,
  2020, pp. 4604--4614.

\bibitem{dropnas}
W.~Hong, G.~Li, W.~Zhang, R.~Tang, Y.~Wang, Z.~Li, and Y.~Yu, ``Dropnas:
  Grouped operation dropout for differentiable architecture search.'' in
  \emph{IJCAI}, 2020, pp. 2326--2332.

\bibitem{foret2020sharpness}
P.~Foret, A.~Kleiner, H.~Mobahi, and B.~Neyshabur, ``Sharpness-aware
  minimization for efficiently improving generalization,'' in
  \emph{International Conference on Learning Representations}, 2020.

\bibitem{tenas}
W.~Chen, X.~Gong, and Z.~Wang, ``Neural architecture search on imagenet in four
  gpu hours: A theoretically inspired perspective,'' in \emph{International
  Conference on Learning Representations (ICLR)}, 2021.

\bibitem{freenas}
M.~Zhang, S.~Su, S.~Pan, X.~Chang, W.~Huang, and G.~Haffari, ``Differentiable
  architecture search without training nor labels: A pruning perspective,''
  \emph{arXiv preprint arXiv:2106.11542}, 2021.

\bibitem{shu2021nasi}
Y.~Shu, S.~Cai, Z.~Dai, B.~C. Ooi, and B.~K.~H. Low, ``Nasi: Label-and
  data-agnostic neural architecture search at initialization,'' in
  \emph{International Conference on Learning Representations}, 2021.

\bibitem{zero}
L.~Xiang, {\L}.~Dudziak, M.~S. Abdelfattah, T.~Chau, N.~D. Lane, and H.~Wen,
  ``Zero-cost proxies meet differentiable architecture search,'' \emph{arXiv
  preprint arXiv:2106.06799}, 2021.

\bibitem{rlnas}
X.~Zhang, P.~Hou, X.~Zhang, and J.~Sun, ``Neural architecture search with
  random labels,'' in \emph{Proceedings of the IEEE/CVF Conference on Computer
  Vision and Pattern Recognition}, 2021, pp. 10\,907--10\,916.

\bibitem{zoph2018learning}
B.~Zoph, V.~Vasudevan, J.~Shlens, and Q.~V. Le, ``Learning transferable
  architectures for scalable image recognition,'' in \emph{Proceedings of the
  IEEE conference on computer vision and pattern recognition}, 2018, pp.
  8697--8710.

\bibitem{dwd}
I.~Loshchilov and F.~Hutter, ``Decoupled weight decay regularization,''
  \emph{arXiv preprint arXiv:1711.05101}, 2017.

\bibitem{hanson1988comparing}
S.~Hanson and L.~Pratt, ``Comparing biases for minimal network construction
  with back-propagation,'' \emph{Advances in neural information processing
  systems}, vol.~1, pp. 177--185, 1988.

\bibitem{prdarts}
P.~Zhou, C.~Xiong, R.~Socher, and S.~C. Hoi, ``Theory-inspired path-regularized
  differential network architecture search,'' in \emph{Proceedings of the 34th
  International Conference on Neural Information Processing Systems}, 2020, pp.
  8296--8307.

\bibitem{neyshabur2017exploring}
B.~Neyshabur, S.~Bhojanapalli, D.~McAllester, and N.~Srebro, ``Exploring
  generalization in deep learning,'' \emph{arXiv preprint arXiv:1706.08947},
  2017.

\bibitem{finlay2018lipschitz}
C.~Finlay, J.~Calder, B.~Abbasi, and A.~Oberman, ``Lipschitz regularized deep
  neural networks generalize and are adversarially robust,'' \emph{arXiv
  preprint arXiv:1808.09540}, 2018.

\bibitem{dong2020bench}
X.~Dong and Y.~Yang, ``Nas-bench-201: Extending the scope of reproducible
  neural architecture search,'' \emph{arXiv preprint arXiv:2001.00326}, 2020.

\bibitem{GDAS}
{X. Dong and Y. Yang}, ``Searching for a robust neural architecture in four gpu
  hours,'' in \emph{Proceedings of the IEEE/CVF Conference on Computer Vision
  and Pattern Recognition}, 2019, pp. 1761--1770.

\bibitem{dsnas}
S.~Hu, S.~Xie, H.~Zheng, C.~Liu, J.~Shi, X.~Liu, and D.~Lin, ``Dsnas: Direct
  neural architecture search without parameter retraining,'' in
  \emph{Proceedings of the IEEE/CVF Conference on Computer Vision and Pattern
  Recognition}, 2020, pp. 12\,084--12\,092.

\bibitem{idarts}
M.~Zhang, S.~W. Su, S.~Pan, X.~Chang, E.~M. Abbasnejad, and R.~Haffari,
  ``idarts: Differentiable architecture search with stochastic implicit
  gradients,'' in \emph{International Conference on Machine Learning}.\hskip
  1em plus 0.5em minus 0.4em\relax PMLR, 2021, pp. 12\,557--12\,566.

\bibitem{CDARTS}
H.~Yu, H.~Peng, Y.~Huang, J.~Fu, H.~Du, L.~Wang, and H.~Ling, ``Cyclic
  differentiable architecture search,'' \emph{IEEE Transactions on Pattern
  Analysis and Machine Intelligence}, 2022.

\bibitem{darts+pt}
R.~Wang, M.~Cheng, X.~Chen, X.~Tang, and C.-J. Hsieh, ``Rethinking architecture
  selection in differentiable nas,'' \emph{arXiv preprint arXiv:2108.04392},
  2021.

\bibitem{wu2019fbnet}
B.~Wu, X.~Dai, P.~Zhang, Y.~Wang, F.~Sun, Y.~Wu, Y.~Tian, P.~Vajda, Y.~Jia, and
  K.~Keutzer, ``Fbnet: Hardware-aware efficient convnet design via
  differentiable neural architecture search,'' in \emph{Proceedings of the
  IEEE/CVF Conference on Computer Vision and Pattern Recognition}, 2019, pp.
  10\,734--10\,742.

\bibitem{chu2021fairnas}
X.~Chu, B.~Zhang, and R.~Xu, ``Fairnas: Rethinking evaluation fairness of
  weight sharing neural architecture search,'' in \emph{Proceedings of the
  IEEE/CVF International Conference on Computer Vision}, 2021, pp.
  12\,239--12\,248.

\bibitem{chu2021scarlet}
X.~Chu, B.~Zhang, Q.~Li, R.~Xu, and X.~Li, ``Scarlet-nas: bridging the gap
  between stability and scalability in weight-sharing neural architecture
  search,'' in \emph{Proceedings of the IEEE/CVF International Conference on
  Computer Vision}, 2021, pp. 317--325.

\bibitem{cai2018proxylessnas}
H.~Cai, L.~Zhu, and S.~Han, ``Proxylessnas: Direct neural architecture search
  on target task and hardware,'' in \emph{International Conference on Learning
  Representations}, 2018.

\bibitem{koonce2021mobilenetv3}
B.~Koonce, ``Mobilenetv3,'' in \emph{Convolutional Neural Networks with Swift
  for Tensorflow}.\hskip 1em plus 0.5em minus 0.4em\relax Springer, 2021, pp.
  125--144.

\bibitem{chu2020moga}
X.~Chu, B.~Zhang, and R.~Xu, ``Moga: Searching beyond mobilenetv3,'' in
  \emph{ICASSP 2020-2020 IEEE International Conference on Acoustics, Speech and
  Signal Processing (ICASSP)}.\hskip 1em plus 0.5em minus 0.4em\relax IEEE,
  2020, pp. 4042--4046.

\bibitem{cai2019once}
H.~Cai, C.~Gan, T.~Wang, Z.~Zhang, and S.~Han, ``Once-for-all: Train one
  network and specialize it for efficient deployment,'' in \emph{International
  Conference on Learning Representations}, 2019.

\bibitem{li2020block}
C.~Li, J.~Peng, L.~Yuan, G.~Wang, X.~Liang, L.~Lin, and X.~Chang,
  ``Block-wisely supervised neural architecture search with knowledge
  distillation,'' in \emph{Proceedings of the IEEE/CVF Conference on Computer
  Vision and Pattern Recognition}, 2020, pp. 1989--1998.

\bibitem{tan2019mixconv}
M.~Tan and Q.~V. Le, ``Mixconv: Mixed depthwise convolutional kernels,''
  \emph{arXiv preprint arXiv:1907.09595}, 2019.

\bibitem{tan2019efficientnet}
M.~Tan and Q.~Le, ``Efficientnet: Rethinking model scaling for convolutional
  neural networks,'' in \emph{International conference on machine
  learning}.\hskip 1em plus 0.5em minus 0.4em\relax PMLR, 2019, pp. 6105--6114.

\bibitem{chu2020noisy}
X.~Chu and B.~Zhang, ``Noisy differentiable architecture search,'' \emph{arXiv
  preprint arXiv:2005.03566}, 2020.

\bibitem{sandler2018mobilenetv2}
M.~Sandler, A.~Howard, M.~Zhu, A.~Zhmoginov, and L.-C. Chen, ``Mobilenetv2:
  Inverted residuals and linear bottlenecks,'' in \emph{Proceedings of the IEEE
  conference on computer vision and pattern recognition}, 2018, pp. 4510--4520.

\bibitem{1shot1}
A.~Zela, J.~Siems, and F.~Hutter, ``Nas-bench-1shot1: Benchmarking and
  dissecting one-shot neural architecture search,'' in \emph{International
  Conference on Learning Representations}, 2019.

\bibitem{panda2021nastransfer}
R.~Panda, M.~Merler, M.~S. Jaiswal, H.~Wu, K.~Ramakrishnan, U.~Finkler,
  C.-F.~R. Chen, M.~Cho, R.~Feris, D.~Kung \emph{et~al.}, ``Nastransfer:
  Analyzing architecture transferability in large scale neural architecture
  search,'' in \emph{Proceedings of the AAAI Conference on Artificial
  Intelligence}, vol.~35, no.~10, 2021, pp. 9294--9302.

\end{thebibliography}
\end{document}